\documentclass[sn-mathphys,Numbered]{sn-jnl}

\usepackage{graphicx}%
\usepackage{multirow}%
\usepackage{amsmath,amssymb,amsfonts}%
\usepackage{amsthm}%
\usepackage{mathrsfs}%
\usepackage[title]{appendix}%
\usepackage{xcolor}%
\usepackage{textcomp}%
\usepackage{manyfoot}%
\usepackage{booktabs}%
\usepackage[ruled,linesnumbered]{algorithm2e}
\usepackage{graphicx}
\usepackage{multirow}
\usepackage{float}
\usepackage{bbm}

\theoremstyle{thmstyleone}%
\theoremstyle{thmstyletwo}%

\theoremstyle{thmstylethree}%

\begin{document}

\title[Article Title]{Artificial Immune System of Secure Face Recognition Against Adversarial Attacks}


\author[1]{\fnm{Min} \sur{Ren}}\email{renmin@bnu.edu.cn}
\equalcont{These authors contributed equally to this work.}

\author[2]{\fnm{Yunlong} \sur{Wang}}\email{yunlong.wang@cripac.ia.ac.cn}
\equalcont{These authors contributed equally to this work.}

\author[3]{\fnm{Yuhao} \sur{Zhu}}\email{zhuyuhao@rails.cn}

\author*[1]{\fnm{Yongzhen} \sur{Huang}}\email{huangyongzhen@bnu.edu.cn}

\author [2]{\fnm{Zhenan} \sur{Sun}}\email{znsun@nlpr.ia.ac.cn}

\author [2]{\fnm{Qi} \sur{Li}}\email{qli@nlpr.ia.ac.cn}

\author [2]{\fnm{Tieniu} \sur{Tan}}\email{tnt@nlpr.ia.ac.cn}

\affil[1]{\orgdiv{School of Artificial Intelligence}, \orgname{Beijing Normal University}, \orgaddress{\city{Beijing}, \country{China}}}

\affil[2]{\orgdiv{State Key Laboratory of Multimodal Artificial Intelligence Systems (MAIS)}, \orgname{Institute of Automation, Chinese Academy of Sciences}, \orgaddress{\city{Beijing},  \country{China}}}

\affil[3]{\orgdiv{Postgraduate Department}, \orgname{China Academy of Railway Sciences}, \orgaddress{\city{Beijing},  \country{China}}}



\abstract{
Deep learning-based face recognition models are vulnerable to adversarial attacks.
In contrast to general noises, the presence of imperceptible adversarial noises can lead to catastrophic errors in deep face recognition models.
The primary difference between adversarial noise and general noise lies in its specificity.
Adversarial attack methods give rise to noises tailored to the characteristics of the individual image and recognition model at hand.
Diverse samples and recognition models can engender specific adversarial noise patterns, which pose significant challenges for adversarial defense.
%
Addressing this challenge in the realm of face recognition presents a more formidable endeavor due to the inherent nature of face recognition as an open set task.
In order to tackle this challenge, it is imperative to employ customized processing for each individual input sample.
Drawing inspiration from the biological immune system, which can identify and respond to various threats, this paper aims to create an artificial immune system (AIS) to provide adversarial defense for face recognition.
%
%
The proposed defense model incorporates the principles of antibody cloning, mutation, selection, and memory mechanisms to generate a distinct ``antibody" for each input sample, wherein the term ``antibody" refers to a specialized noise removal manner.
Furthermore, we introduce a self-supervised adversarial training mechanism that serves as a simulated rehearsal of immune system invasions.
%
%
Extensive experimental results demonstrate the efficacy of the proposed method, surpassing state-of-the-art adversarial defense methods.
The source code is available \href{https://github.com/RenMin1991/SIDE}{here}, or you can visit this website: \textit{https://github.com/RenMin1991/SIDE}

}




\maketitle


\section{Introduction}

Deep learning-based feature extractors have garnered significant achievements across diverse domains, notably in image classification~\cite{lecun1998gradient, Krizhevsky2012ImageNet, Simonyan2014Very, Szegedy2015Going, Huang2017Densely, Hu2018Squeeze, He2016Deep}, object detection~\cite{ren2015faster, redmon2016you}, and semantic segmentation~\cite{long2015fully, he2017mask}.
This can be attributed to their capability of executing non-linear mappings from raw data to high-dimension features.
Despite the powerful expressive capabilities of deep learning models, their susceptibility to adversarial attacks undermines their reliability and degrades their security~\cite{Szegedy2013Intriguing, Goodfellow2014Explaining, S2017Universal, wei2023infrared, zheng2023u}. 
Hence, numerous researchers have directed their attention towards adversarial defense techniques for deep learning models~\cite{Kurakin2016Adversarial, Na2018Cascade, Tram2018Ensemble, Ross2017Improving, shao2022open, dolatabadi2023adversarial, zhang2023comprehensive, liu2023towards}.

As a prevalent and widely adopted application of deep learning technology, deep learning based face recognition~\cite{Taigman2014DeepFace, Yi2014Deep, Schroff2015FaceNet, Liu2017SphereFace, Wang2018CosFace, Deng2018ArcFace, ren2020dynamic, ren2023Multiscale, Joint2023Zhu} have demonstrated their ability to surpass human performance in both verification and identification scenarios.
The domains where face recognition is applied, such as finance and border control, typically impose stringent requirements on security.
Nevertheless, the security of deep learning-based face recognition systems is greatly degraded by the inherent fragility of deep learning models against adversarial attacks.
Whether these attacks stem from the digital domain~\cite{Dong2019Efficient} or are directly imposed in the physical domain~\cite{Komkov2013AdvHat}, they effortlessly exploit vulnerabilities within face recognition models, leading to catastrophic errors.
Thus, the pursuit of adversarial defense methods in the domain of face recognition not only holds theoretical significance but also represents a pressing technological imperative driven by practical application demands.

In fact, deep learning models demonstrate remarkable robustness against common types of noise, such as Gaussian and salt-and-pepper noise.
Their susceptibility to adversarial noises can be attributed to the specificity possessed by these noises.
Adversarial attack methods generate noises that are specifically tailored to the characteristics of the target image and recognition model.
As a consequence, distinct samples and recognition models can lead to the creation of specific patterns of adversarial noise, thus imposing considerable challenges when devising effective mechanisms for adversarial defense.


Facial recognition can be applied to both close-set scenarios, where all identities are available, and open-set scenarios, where the identities encountered during testing are unavailable.
This challenge poses great difficulties for the face recognition task due to the fact that face recognition tasks involve dealing with a substantially larger number of unique identities, especially in open-set scenarios. 
%
Hence, adversarial samples in the context of face recognition demonstrate a great degree of diversity, complexity, and unpredictability in their adversarial noise patterns.
The specificity inherent in adversarial noises amplifies their detrimental impact, thereby intensifying the vulnerability of face recognition models to adversarial attacks.

   %
   %
   %

In response to this challenge, we propose a novel adversarial defense model for face recognition.
The proposed model provides individualized manners for removing adversarial noises, thus endowing each facial image with a tailored and specific defense against adversarial noises.
This approach draws inspiration from the biological immune system, which exhibits powerful attributes such as self-learning, dynamic adaptation, and memory capabilities~\cite{Burnet1957A}.
The immune system can generate specific antibodies against various viruses through processes of antibody cloning, mutation, selection, and memory.
These mechanisms enable the production of effective antibodies that bind specifically to antigens, thus achieving effective immunity.

In the proposed method, adversarial noises can be analogous to antigens, while the noise removal ways are analogous to antibodies.
%
When devising the structure of antibodies, the perturbation inactivation methodology of PIN~\cite{Ren2022Perturbation} is assimilated.
This method is an adversarial defense approach that utilizes eigenvectors of facial images to filter noise and restoring essential facial features.
%
However, PIN fails to consider the specificity of adversarial noises during noise removal, making it challenging to effectively differentiate between the harmful information introduced by adversarial noises and the inherent information of the face itself.
As a result, striking a balance between adversarial noise removal and facial sample restoration becomes difficult.
This issue is also encountered by most adversarial defense methods based on noise removal.
%
In contrast, we propose an artificial immune system that provides customized noise removal ways for facial samples.
This is achieved through the generation, cloning, mutation, and memory mechanisms of antibodies within the immune system.
The proposed defense model consists of three essential components: the antigen analyzer, the antibody generator, and the memory module.
The antigen analyzer is responsible for analyzing the characteristics of adversarial noises, whereas the antibody generator emulates the processes of antibody cloning, mutation, and selection to optimize the antibodies.
Concurrently, the memory module is employed to store patterns of adversarial noises during the optimization process.
%

%


In addition, we propose a self-supervised adversarial training mechanism that collaborates with the aforementioned adversarial defense model.
This mechanism integrates a momentum-updated siamese network of the adversarial defense model to generate on-the-fly adversarial samples.
%
Self-supervised adversarial training can offer more precise guidance for the process of antibody selection, leading to a gradual enhancement in their defense capabilities.
This process is analogous to how the immune system continuously enhances its immune capabilities through repeated confrontations with viral invasions.

The main contributions of this paper can be summarised as follows:

\begin{itemize}

\item In this paper, we introduce a novel face recognition adversarial defense model based on the principles of specific immunity.
By emulating the intricate process of specific immune evolution observed in biological systems, this model provides tailored denoising ways for individual input facial images.
This model effectively mitigates the challenges posed by the specificity of adversarial noise to face recognition models, enabling robust and reliable recognition performance.

\item We introduce a self-supervised adversarial training mechanism that contributes to the selection of ``antibodies" within the proposed adversarial defense model.
This mechanism facilitates iterative refinement through self-adversarial training, empowering the model to enhance its defensive capabilities against adversarial noises.

\item  The effectiveness of the proposed adversarial defense method has been experimentally validated across diverse types of adversarial attacks and multiple datasets.
The experimental results demonstrate its superior performance compared to existing state-of-the-art adversarial defense methods.
Furthermore, we offer a comprehensive experimental analysis, providing valuable insights into the effectiveness and robustness of the proposed method.

\end{itemize}

The remainder of this paper is organized as follows:
Section~\ref{sec:RelateWork} presents a brief literature review of the related work.
The proposed adversarial defense model and the self-supervised adversarial training mechanism are described in detail in Section~\ref{sec:Method}.
The configurations and results of experiments are presented in Section~\ref{sec:Experiments}.
Finally, the conclusion of this paper is summarized in Section~\ref{sec:Conclusion}.


\section{Related Work}
\label{sec:RelateWork}

\subsection{Deep Learning Based Face Recognition}

Deep learning-based face recognition has achieved remarkable advancements in recent years.
The pioneering work of CNN-based face representation was introduced by Taigman et al.~\cite{Taigman2014DeepFace} and Yi et al.~\cite{Yi2014Deep}.
In these studies, face recognition is regarded as a multi-class classification challenge.
To address this, deep convolutional neural network (CNN) models are initially implemented to acquire features from extensive datasets containing multiple identities.
However, due to the fact that face recognition is an open-set task, meaning testing identities that are different from the ones used for training, there is a significant difference between face recognition and image classification tasks.
As a result, researchers have increasingly tended to model it as a metric learning task, aiming to learn a feature space mapping model with strong discriminative power through constraints imposed on the feature space.
In the pursuit of obtaining a 128-D face embedding representation, Schroff et al. utilized a triplet loss function in their research~\cite{Schroff2015FaceNet}.
To further enhance feature embedding, numerous methods have been proposed, such as SphereFace~\cite{Liu2017SphereFace}, CosFace~\cite{Wang2018CosFace}, ArcFace~\cite{Deng2018ArcFace}, among others.
In addition, to facilitate the deployment of facial recognition models, researchers have also focused on the study of lightweight face recognition models~\cite{wu2018light, Duong2019Mobiface}.

In spite of the considerable progress made in this field, deep learning models utilized for face recognition remain susceptible to adversarial attacks, as indicated by previous studies~\cite{Dong2019Efficient, Komkov2013AdvHat, zhong2020toward}.
This vulnerability presents a profound concern and jeopardizes the overall security of face recognition systems.


\subsection{Adversarial Attack}

The adversarial attack technique for computer vision tasks has become a prominent area of research.
Szegedy et al.~\cite{Szegedy2013Intriguing} is the first to demonstrate the vulnerability of deep neural networks to adversarial noises.
Since then, numerous methods for adversarial attacks have been proposed.
Goodfellow et al.~\cite{Goodfellow2014Explaining} introduced an efficient single-step attack method called FGSM, which is based on gradient calculations.
DeepFool~\cite{Moosavi2016DeepFool} aims to identify the nearest decision boundary in order to confuse the model.
C\&W~\cite{Carlini2017Towards} addresses the joint optimization of the objective function and the scale of noises.
Projected gradient descent (PGD)~\cite{Madry2017Towards} iteratively applies the gradient signal of deep learning models, which is the most powerful first-order adversarial attack method~\cite{Madry2017Towards}.
Su et al.~\cite{su2019one} propose an intriguing approach that confuses deep learning models by altering just a single pixel in the image.
Additionally, there have been reports on the generalization of adversarial noises~\cite{Zhang2023Improving, Zhang2023Transferable, Liu_2023_CVPR, Wang_2023_CVPR, Liang_2023_CVPR}.
Universal adversarial noises based attack methods are proposed in several studies~\cite{S2017Universal, zhang2021data, wang2021feature, yuan2021meta}.

Recently, there have been reports of targeted adversarial attack techniques specifically tailored for face recognition systems.
Dong et al.~\cite{Dong2019Efficient} introduce a decision-based adversarial attack method for face recognition.
The rapidly evolving field of transferable facial adversarial attack techniques has provided additional avenues for black-box facial adversarial attacks~\cite{Li2023Sibling, Li2023Discrete}.
Furthermore, instances of physical domain adversarial attacks have also been documented in the literature.
Sharif et al.~\cite{Sharif2016Accessorize} presented a systematic approach for generating physically feasible attacks by printing a pair of eyeglass frames.
Another method, known as sticker attack, was proposed by Komkov et al.~\cite{Komkov2013AdvHat}, which involves using a specially designed rectangular paper sticker to deceive the face recognition system.
Recently, there have been further developments in sticker-based adversarial attacks on faces. 
Yang et al.~\cite{Yang_2023_CVPR} introduce a sticker generation method that can adhere to the three-dimensional shape of faces.
These real-world attacks pose new challenges to the adversarial defense for face recognition systems.


\subsection{Adversarial Defense}

The extant adversarial defense methods can be broadly categorized into two distinct groups.
The first group encompasses methods that strive to enhance the robustness of neural networks against adversarial examples.
These methods concentrate on fortifying the network's capacity to withstand noises and maintain accurate predictions in the presence of such malicious input.
On the other hand, the second group involves methodologies that aim to eliminate the adversarial noises from the adversarial samples prior to their presentation to the recognition model.
This category of approaches focuses on cleansing the input data by removing the embedded harmful alterations, thereby reducing the potential impact of adversarial attacks.

A prevalent strategy of the first type involves training neural networks using adversarial examples~\cite{Kurakin2016Adversarial, Goodfellow2014Explaining, Na2018Cascade, Tram2018Ensemble, Hsiung_2023_CVPR}.
This strategy is straightforward and aims to enhance the network's resistance against adversarial attacks.
%
%
To improve the robustness against gradient-based attacks, several learning strategies have been proposed.
Ross et al.~\cite{Ross2017Improving} trained models with input gradient regularization.
Other techniques, such as network distillation~\cite{papernot2016distillation}, region-based classifier~\cite{cao2017mitigating}, generative model~\cite{lee2017generative, jang2019adversarial}, and self-supervised learning~\cite{moayeri2021sample} have also been adopted to enhance model robustness.
Rakin et al.~\cite{he2019parametric} introduced a trainable randomness method for adversarial training to improve robustness.
A novel loss function for adversarial defense is proposed for adversarial defense in~\cite{chen2019improving}.
Mustafa et al.~\cite{mustafa2019adversarial} achieved elevated robustness by constraining the hidden space of deep neural networks.
Zhong et al.~\cite{zhong2019adversarial} utilized margin-based triplet embedding regularization to train the recognition model.
Cazenavette et al.~\cite{cazenavette2021architectural} aimed to enhance the adversarial robustness of CNNs by reframing each layer as a sparse coding model. 
Jin et al.~\cite{Jin_2023_CVPR} present an approach for analyzing the noise pattern by Taylor expansion.
However, these methods often exhibit poor generalization to adversarial noise that does not appear in the training set, as verified by our experiments in this paper.
This is because the patterns of adversarial noise are more complex and diverse in face recognition tasks, and relying solely on adversarial training is insufficient to cope with them.

The other type of approaches are devised to eliminate the adversarial noises prior to the recognition model's processing~\cite{das2017keeping, guo2017countering, Ren2022Perturbation, meng2017magnet, liao2018defense, song2017pixeldefend}.
Das et al.~\cite{das2017keeping} proposed the application of JPEG compression to remove these noises.
In a study by Guo et al.~\cite{guo2017countering}, image quilting and total variation minimization (TVM) were assessed as possible techniques for this purpose.
Meng et al.~\cite{meng2017magnet} introduced a two-pronged defense strategy to effectively eliminate adversarial noises. 
Liao et al.~\cite{liao2018defense} incorporated the U-Net~\cite{ronneberger2015u} as a denoising module, thereby enabling the removal of adversarial noises.
The work presented in~\cite{song2017pixeldefend} employed PixelCNN~\cite{song2017pixeldefend} to transform adversarial examples into clean images.
Bai et al.~\cite{bai2019hilbert} further improved the defense performance by incorporating Hilbert scan into PixelCNN.
Dezfooli et al.~\cite{moosavi2018divide} and Sun et al.~\cite{sun2019adversarial} utilized sparse coding to reconstruct image patches.
Gupta et al.~\cite{gupta2019ciidefence} attempted to identify the most influential regions of an image for reconstruction.
Xie et al.~\cite{xie2019feature} employed a self-attention layer to recover the original information within the feature space.
Zhou et al.~\cite{zhou2021removing} utilized self-supervised learning to eliminate adversarial noise in the class activation feature space.
PIN~\cite{Ren2022Perturbation} utilizes eigenvectors of facial images to filter noise and restoring essential facial features.
However, it is worth noting that most of these methods were primarily developed for general image classification tasks.
%
%
Furthermore, they are unable to provide specific noise removal ways, rendering them unsuitable for face recognition tasks.

%


\subsection{Algorithms Inspired by the Immune System}

The biological immune system is an evolved defense mechanism in vertebrates to protect the organism from the invasion of ``non-self" entities such as pathogens.
Due to its superior characteristics, including self-learning, memory mechanisms, and dynamic adaptability~\cite{Burnet1957A}, the biological immune system has provided abundant biomimetic inspiration for solving various problems.
Artificial immune systems (AIS)~\cite{Castro2002Artificial} construct algorithms by simulating the functions, principles, and models of the biological immune system.

Among numerous immunological theories, the clonal selection theory~\cite{Burnet1957A} has provided significant inspiration for the development of computer algorithms~\cite{Castro2002Artificial, Chandrasekaran2006Solving, Cutello2007An, Luo2019A, Jie2008Clonal}.
The clonal selection theory explains the fundamental characteristics of adaptive immune response under antigen stimulation.
Its basic concept is that only those cells capable of recognizing antigens are selected for proliferation, while those incapable of antigen recognition are not selected.
The selected cells undergo proliferation and mutation processes to enhance their affinity.

Drawing inspiration from the principles of the clonal selection theory, researchers have put forth numerous algorithms that have yielded fruitful research outcomes in domains such as dynamic programming~\cite{Luo2019A} and multi-objective optimization~\cite{Jie2008Clonal}.
Clonal selection algorithms commonly incorporate the following essential components: 
affinity calculation, which quantifies the quality of antibodies; 
selection, which involves the screening of existing antibodies; 
cloning, which encompasses the replication of antibodies; 
mutation, which entails modifying the structure of antibodies; 
and memory, which involves the storage of information regarding antibodies~\cite{Castro2002Artificial}.

This paper stands as a pioneering effort that amalgamates the clonal selection algorithm with deep learning methodologies for adversarial defense.
We propose a novel approach to address the specificity and complexity of adversarial noise in face recognition tasks by implementing the aforementioned components of the clonal selection algorithm.
This approach has successfully achieved adversarial defense tailored for face recognition.


\section{Methodology}
\label{sec:Method}


This section provides a comprehensive description of the proposed defense method.
Firstly, we present the preliminaries of the method, encompassing symbol representation and the definition of antibodies.
Building upon this foundation, we introduce the structure and overview of the proposed model.
Subsequently, we delve into the defense model optimization and the self-supervised adversarial defense training.
Finally, we provide the implementation details of the method.

\subsection{Preliminaries}
\label{sec:prelim}

A face recognition model can be regarded as a mapping function that transforms facial images into the feature space.
Given a facial image represented as $x \in \mathbb{R}^{C \times H \times W}$, the deep feature $f$ can be extracted using a face recognition model denoted as $F$:
\begin{equation}
    f = F(x)
\end{equation}
%
Adversarial samples are generated by the adversarial attack method according to the facial image $x$ and the face recognition model $F$:
\begin{equation}
    x_{adv} = Attacker(x, F)
\end{equation}
where $Attacker$ is the adversarial attack method, $x_{adv}$ is the adversarial sample.
The adversarial noise is the difference between $x$ and $x_{adv}$.
In this paper, we propose an adversarial defense model, denoted as $D$, in this paper for recovering $x$ from $x_{adv}$:
\begin{equation}
    x_{recon} = D_\theta(x_{adv})
\end{equation}
where $\theta$ is the parameters of the proposed model.

\paragraph{Definition of Antibody:}
In the context of adversarial defense, the adversarial noises in adversarial samples can be viewed as antigens, while noise removal methods can be seen as antibodies.
%
The fundamental motivation behind the design of antibody form lies in facilitating the analysis and removal of noise.
Given the intricate and diverse nature of adversarial noises in facial recognition, modeling the distribution of clean facial images in pixel space serves as a foundation for mitigating the challenges associated with noise analysis.
By acquiring a comprehensive understanding of the underlying distribution of clean facial images, we can establish a solid basis for reducing the complexity involved in noise analysis and consequently lowering the difficulty in effectively removing adversarial noises.


In the realm of facial image distribution modeling, researchers have already made significant strides, providing valuable insights for further investigation.
Among these noteworthy contributions, EigenFace~\cite{turk1991face} stands out as a pioneering and groundbreaking approach.
%
Following the lead of EigenFace, we can model the distribution of facial images in pixel space by the eigenvectors that characterize the data distribution.
Each eigenvector represents a distinct dimension in pixel space, and the components of facial images across different eigenvectors express various types of features.
%
This provides a favorable foundation for us to design noise removal methods, namely antibodies.
%
Therefore, we define antibodies using eigenvectors: an antibody is a composition of eigenvectors: 
\begin{equation}
   a = \{ e_1, e_2, ..., e_n \}
\end{equation}
where $a$ is an antibody, $e_i\in \mathbb{R}^d, ~ d=C \times H \times W$ is an eigenvector of the facial images, $n$ is the number of eigenvectors of the antibody, and $0<n<d$.


The eigenvectors comprising antibodies have the ability to selectively filter features of a facial image, retaining the characteristics corresponding to these eigenvectors:
\begin{equation}
    \alpha = E^T(x^{flat}_{adv}-x_{mean})
\end{equation}
where $E \in \mathbb{R}^{d \times n}$ is a matrix composed of the eigenvectors from an antibody, wherein each column represents an eigenvector, $x^{flat}_{adv} \in \mathbb{R}^d$ is obtained by reshaping $x_{adv}$ into a flattened vector, $x_{mean} \in \mathbb{R}^d$ is the mean vector of the facial samples in the pixel space.
%
$\alpha \in \mathbb{R}^n$ represents the component of the input image along the eigenvectors of the antibody, encompassing only the characteristic information associated with these eigenvectors. 
%
After obtaining $\alpha$, the facial image can be reconstructed:
\begin{equation}
    x_{recon} = E\alpha^T + x_{mean}
\end{equation}


%
By performing the aforementioned processing steps, the reconstructed image will only contain the features corresponding to the eigenvectors of the antibody, while the features corresponding to eigenvectors not present in the antibody are removed.
For different adversarial samples, employing distinct antibodies enables the removal of different features, thereby facilitating a tailored and specific noise removal manner for each input sample.
%


\begin{figure}[t]
\begin{center}
\includegraphics[width=0.9\linewidth]{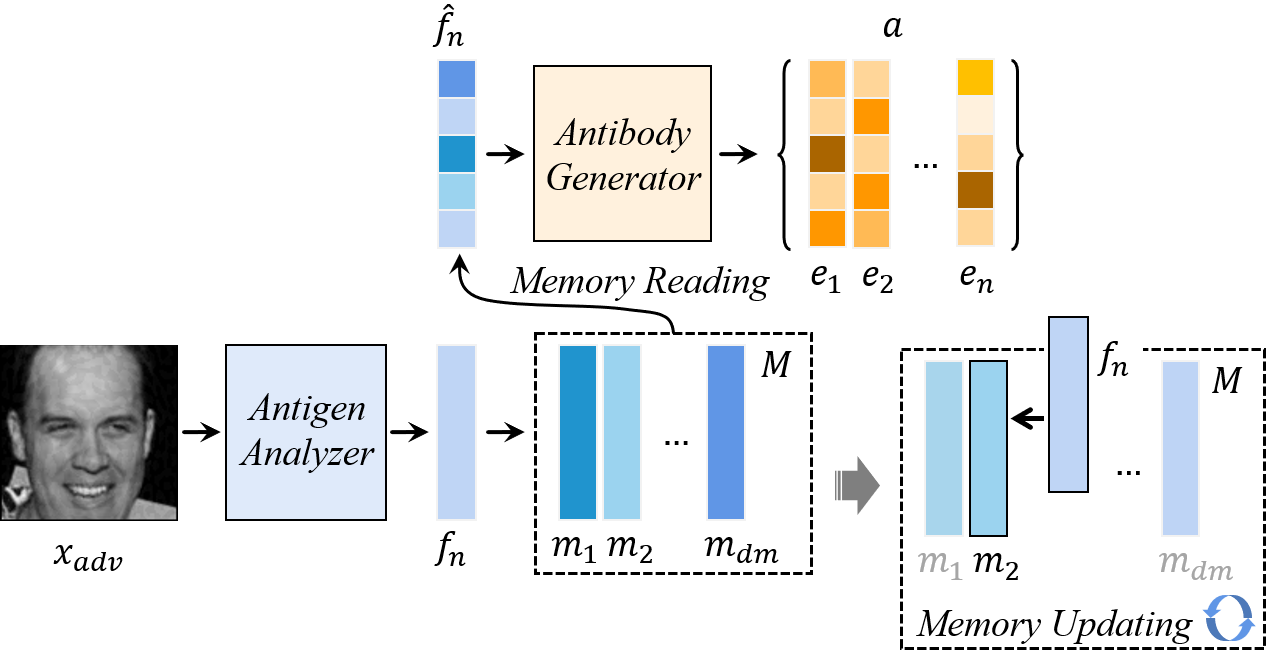}
\end{center}
   \caption{The architecture of the proposed adversarial defense method. The proposed adversarial defense model encompasses three key components: the antigen analyzer, the antibody generator, and the memory module.}
\label{fig:framework}
\end{figure}

\begin{figure}[t]
\begin{center}
\includegraphics[width=0.8\linewidth]{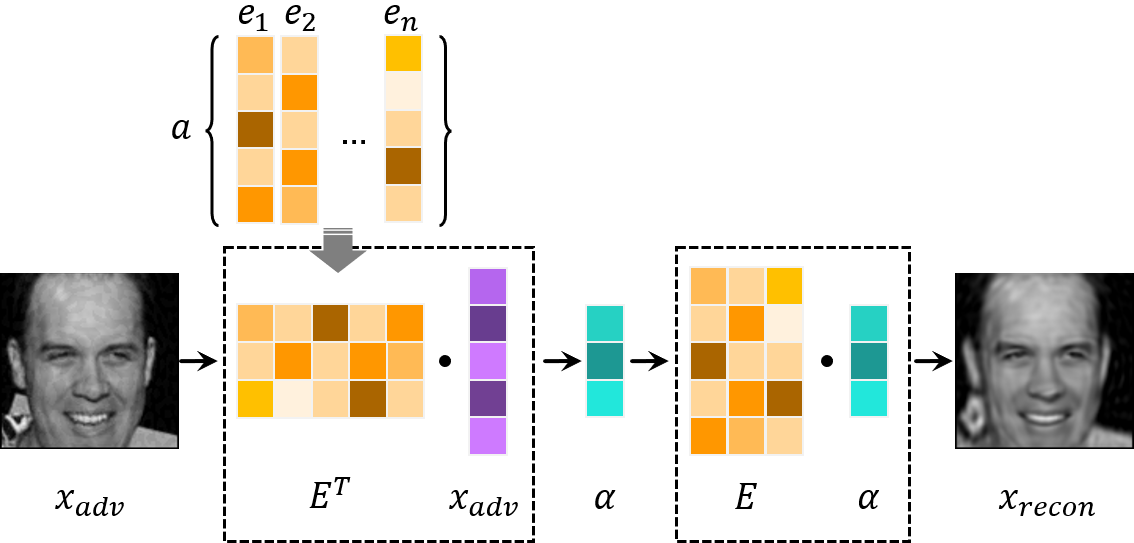}
\end{center}
   \caption{The process of using antibodies for noise removal (omitting $x_{mean}$ for brevity). The eigenvectors comprising antibodies have the ability to selectively filter facial features in an image, retaining the characteristics corresponding to these eigenvectors while removing the remaining information.}
\label{fig:recon}
\end{figure}



\subsection{Model Architecture}
\label{sec:arc}

%
The proposed adversarial defense model encompasses three key components: the antigen analyzer, the antibody generator, and the memory module, as shown in Fig.~\ref{fig:framework}.


In order to provide specific treatment for each input adversarial face image, it is essential to conduct an effective analysis of these images.
To this end, we adopt a deep neural network, serving as a noise analyzer for adversarial samples, namely the antigen analyzer.
%
Given an adversarial sample $x_{adv}$, the antigen analyzer, denoted as $H$, takes the adversarial sample as its input and produces the noise feature $f_n \in \mathbb{R}^{d_n}$ as output:
\begin{equation}
\label{eq:H}
    f_n = H(x_{adv})
\end{equation}
%
The noise information of the input image is contained in $f_n$.


Inspired by the memory mechanisms in the immune system, a memory module is incorporated into the proposed defense model.
This memory module is used to store the noise patterns of adversarial samples, enabling the defense model to explicitly model vulnerabilities in the face recognition model and guide the generation of ``antibodies".
%
%
%
%
The memory module, represented by the noise feature matrix $M \in \mathbb{R}^{d_n \times d_m}$, serves as a repository of noise patterns, with $d_m$ denoting the number of memory items.
Before generating antibodies, $f_n$ is utilized to retrieve noise features from the memory module $M$.
Subsequently, a self-attention mechanism is employed to aggregate the retrieved noise features as the input of antibody generator $\hat{f}_n$.
%
The detailed process of feature retrieval and aggregation, as well as the updating of the memory module, will be described in Section~\ref{sec:memory}.

The antibody generator initially maps the aggregated noise feature $\hat{f}_n$ onto the eigenvector selection space through a feature mapping layer:
\begin{equation}
\label{eq:G}
    f_e = Sigmoid(G(\hat{f}_n))
\end{equation}
where $G$ is the mapping layer, Sigmoid function is employed to normalize the output within the range of $(0, 1)$.
%
The dimension of $f_e$ is the same as the number of eigenvectors, with each dimension corresponding to the probability of including a specific eigenvector in the antibody.
Subsequently, an antibody, which is a combination of eigenvectors, is obtained by sampling based on $f_e$.
After obtaining the antibody, a tailored and specific noise removal manner for each input sample can be implemented as described in Section~\ref{sec:prelim}.

\subsection{Memory Mechanism}
\label{sec:memory}

%
In the immune system, the memory mechanism enable the retention of antigenic information, empowering it to swiftly generate effective antibodies upon encountering similar antigens.
Taking inspiration from this phenomenon, we leverage a memory module to store the noisy features captured during the training process.


\paragraph{Memory Reading:}
%
Upon obtaining the noise feature $f_n$, we employ it as a query to retrieve the items in the memory module $M \in \mathbb{R}^{d_n \times d_m}$:
 \begin{equation}
 \label{eq:memory}
     r_i = \frac{m_if^T_n}{||m_i||~||f_n||}, ~i=1,2,...,d_m
 \end{equation}
where $m_i \in \mathbb{R}^{d_n}$ is the $i$-th item of $M$, $d_m$ is the number of memory items, $r_i$ is the similarity between the query $f_n$ and the $i$-th item.
%
Subsequently, by aggregating the memory items by soft-attention, we can accomplish memory retrieval and obtain the output of the memory module:
\begin{equation}
    \hat{f}_n = \sum_{i=1}^{d_m} r_im_i
\end{equation}

\paragraph{Memory Updating:}
%
During the training process, continuous updating of the memory model is necessary to adapt to changes in model parameters.
To achieve this, we first identify the memory item that is most similar to $f_n$:
\begin{equation}
    i^* = \mathop{\arg\max}\limits_i \frac{m_if^T_n}{||m_i||~||f_n||}, ~i=1,2,...,d_m
\end{equation}
%
Afterwards, the memory module is updated via a moving average of $f_n$:
\begin{equation}
\label{eq:MemUp}
    m_i \leftarrow \begin{cases}
        m_i & i\neq i^*;\\
        \epsilon m_i + (1-\epsilon)f_n & i = i^*.
    \end{cases}
\end{equation}
where $\epsilon \in (0, 1)$ is the decay rate of memory updating.
%
This means that the model updates only the memory item that is most similar to $f_n$ each time while leaving the other items unchanged.
As new adversarial noises continue to emerge, the memory module can store previous noise features.


\subsection{Model Optimization}

To train the defense model presented in the previous subsection, this subsection introduces the method for optimizing the model.
%
%
Drawing inspiration from the immune system's generation and selection process of antibodies, we propose an antibody optimization approach that includes the following elements: antibody affinity, cloning and mutation, antibody screening, and model updating.
In this subsection, we introduce each of these elements in detail and then present the antibody optimization algorithm based on them.

\paragraph{Antibody affinity:}
In immunology, antibody affinity refers to the strength of the bond between an antibody and an antigen.
The higher the affinity, the better the antibody's ability to neutralize the antigen and the more effective the immune response.
We borrow this concept to measure the effectiveness of noise removal in adversarial defense.
%
The definition of antibody affinity is as follows:
\begin{equation}
\label{eq:affinity}
    s(a) = \lambda_1 cosine(F(x_{recon}), F(x)) - \lambda_2 ||x_{recon} - x||_2 - \lambda_3 |a|
\end{equation}
where $cosine(\cdot, \cdot)$ refers to cosine similarity, $F(\cdot)$ refers to the face recognition model, $\lambda_1, \lambda_2, \lambda_3$ are the weights of losses.
%
The antibody affinity includes three items, the first two items are:
the cosine similarity between the denoised facial image and the original image in the deep feature space; the Euclidean distance between them in the pixel space.
These two items measure the difference between the denoised image and the original clean image in both feature space and pixel space, which directly reflects the effectiveness of removing adversarial noise.
The third item is the regularization term of the antibody, which constrains the number of eigenvectors contained in the antibody.
In other words, it is desired that the antibody can extract the most critical features of the facial image with as few eigenvectors as possible.
%

\paragraph{Cloning and Mutation:}
%
%
In the immune system, the cloning and mutation of antibodies are two important mechanisms.
On the one hand, antibody cloning can maintain the characteristics of effective antibodies relatively stably.
On the other hand, an appropriate degree of mutation during the clonal process allows for antibody variation, enabling the immune system to effectively respond to changes in antigens.
It is the effective coordination of these two mechanisms that endow the immune system with strong dynamic adaptability.
In the task of adversarial defense for face recognition, it is also necessary to balance the stability and dynamic adaptability of antibodies.


Therefore, in the optimization of antibodies, we achieve the cloning and mutation of antibodies by sampling according to $f_e$ in Eq.~\ref{eq:G}.
%
%
As described in Section~\ref{sec:arc}, each component of $f_e$ represents the probability of incorporating an eigenvector into the antibody.
Therefore, during the optimization process, we sample $k$ antibodies based on $f_e$, which serves as the cloning process of the antibodies.
When the sampling probabilities in $f_e$ are close to 0 or 1, the $k$ antibodies obtained by sampling will have strong consistency, and the probability of antibody mutation is small.
Conversely, when the probability within $f_e$ is close to 0.5, the probability of antibody mutation increases, and the differences between the sampled antibodies also become larger.
Through the aforementioned sampling process, we complete the cloning and mutation of antibodies.

\paragraph{Antibody Screening and Model Updating:}
%
After obtaining the sampled antibodies $\{ a_1, a_2, ..., a_k \}$, the defensive ability of each antibody can be measured using Eq.\ref{eq:affinity}: $\{ s(a_1), s(a_2), ..., s(a_k) \}$.
For antibodies with stronger defensive abilities, we hope to increase their probability of being sampled based on $f_e$ in order to obtain more similar antibodies.
Conversely, for antibodies with weaker defensive abilities, we hope to decrease their probability of being sampled, in order to reduce the number of similar antibodies.

To achieve this objective through the updating of model parameters, we first calculate the likelihood of each antibody: $l(a_i)$.
%
The gradient of $l(a_i)$ with respect to the parameters of the proposed model $\theta$ represents the direction of increasing likelihood of $a_i$.
Therefore, we adjust the gradient using antibody affinity to obtain the direction of model updating, and then implement model update through gradient descent:

\begin{equation}
\label{eq:ModelUpdate}
    \theta \leftarrow \theta - \frac{\phi}{k} \sum_{i=1}^k (s_0-s(a_i)) \nabla_\theta l(a_i), ~~ s.t. ~~ s_0 = \frac{1}{k} \sum_{i=1}^k s(a_i)
\end{equation}
where $\phi$ is the learning rate, $s_0$ provides a baseline for the evaluation of antibodies.
The likelihood of antibodies with an affinity higher than $s_0$ will increase, while the likelihood of antibodies with an affinity lower than $s_0$ will decrease.

The pseudo-code for the proposed model optimization algorithm is shown in Algorithm~\ref{alg:antibody}.

\begin{algorithm}
\caption{Model Optimization Algorithm}\label{alg:antibody}

\KwIn{raw facial images $\{ x^i \}$, corresponding adversarial samples $\{ x^i_{adv} \}$, $D_\theta$, number of sampling $k$, learning rate $\phi$}
\BlankLine
\For{$i \leftarrow 1$ \KwTo number of facial images}{

    \emph{get $f^i_e$ by feeding $x^i_{adv}$ into the defense model}\;
    \emph{get $\{ a^i_1, a^i_2, ..., a^i_k \}$ by cloning according to $f^i_e$}\;
    \For{$j \leftarrow 1$ \KwTo $k$}{
        \emph{get the likelihood $l(a^i_j)$}\;
        \emph{get antibody affinity $s(a^i_j)$ according to Eq.\ref{eq:affinity}}\;
    }
    \emph{update $\theta$ accroding to Eq.\ref{eq:ModelUpdate}}
    
}
\end{algorithm}


\subsection{Self-supervised Adversarial Training}

In order to provide effective guidance for model optimization, we propose a self-supervised adversarial training mechanism.
This mechanism involves generating adversarial samples to purposefully train the proposed adversarial defense model.
%
The effectiveness of adversarial training relies on two conditions.
Firstly, generating an ample amount of adversarial samples is crucial to prevent the model from taking shortcuts and ensure its generalization.
Secondly, the generation process of adversarial samples needs to exhibit consistency to maintain training stability.
However, these two aspects are contradictory under the constraint of limited storage space.
While ensuring consistency, the limited storage space restricts the number of adversarial samples involved in training.


To address this contradiction, we draw inspiration from MoCo~\cite{He_2020_CVPR} and employ a siamese model $D_{\Bar{\theta}}$ for the defense model, where its parameters are updated with an exponential moving average of $D_\theta$:
\begin{equation}
\label{eq:AdvTrain}
    \Bar{\theta} \leftarrow \xi \Bar{\theta} + (1-\xi) \theta
\end{equation}
where $\xi \in (0, 1)$ is the decay rate of updating.
As $\xi$ approaches 1, the differences between the adversarial samples from different mini-batches are relatively small, thereby satisfying both the requirement for sample quantity and the demand for consistency.
%
For each facial image $x$, a corresponding adversarial sample can be generated using this siamese model.
When generating adversarial samples, we employ FGSM, which perturbs the facial sample by making small adjustments in the direction of the gradients of the siamese model:
\begin{equation}
\label{eq:FGSM}
    x_{adv} = x + \eta ~ sign(\nabla_x L(x))
\end{equation}
\begin{equation}
\label{eq:loss}
    L(x) = 1-cosine(F(x), F(D_{\Bar{\theta}}(x)))
\end{equation}
where $sign(\cdot)$ refers to sign function, $cosine(\cdot, \cdot)$ refers to cosine similarity, $F(\cdot)$ refers to the face recognition model, $\eta$ is the scale of adversarial noises.


Through the proposed adversarial self-supervised training, we have established a training mechanism that concurrently fulfills the requirements of both adversarial sample quantity and consistency.
This mechanism serves as an effective guidance for the optimization of the adversarial defense model.
%


\subsection{Implementation Details}
\label{sec:detail}
The implementation details of the proposed adversarial defense method are introduced in this subsection.


The antigen analyzer $H$ in Eq.\ref{eq:H} is ResNet-18~\cite{He2016Deep}, which consists of a compendium of 17 convolutional layers complemented by a fully connected layer.
The dimension of $f_n$, which is the output of $H$, is 512.
%
The number of memory items $d_m$ in Eq.\ref{eq:memory} is set to 128.
The feature mapping layer $G$ in Eq.\ref{eq:G} is a fully connected layer, which maps the noise features onto the eigenvector selection space.
%
During the training process, we exclusively consider the top 1500 eigenvectors with the largest eigenvalues and disregard the rest by assigning a probability of zero to their selection.
Consequently, the dimension of the eigenvector selection space is 1500, resulting in the dimension of $f_e$, which is the output of $G$, is 1500.


The proposed adversarial defense model is trained on the CelebA Dataset~\cite{liu2015faceattributes}, which comprises a substantial collection of 202,599 facial images.
To ensure consistency, all training images underwent alignment~\cite{Deng2018ArcFace} and were resized to $112 \times 112$.
Additionally, a transformation is applied to convert them to grayscale, facilitating the computation of eigenvectors.
%
The eigenvectors are obtained on the gray-scale facial images of CelebA Dataset.


The decay rate of memory updating $\epsilon$ in Eq.\ref{eq:MemUp} is set to 0.999.
The loss weights $\lambda_1, \lambda_2, \lambda_3$ in Eq.\ref{eq:affinity} are set to 8, 1, and 0.003, respectively.
The decay rate of siamese model updating $\xi$ in Eq.\ref{eq:AdvTrain} is set to 0.999.
The noise scale during self-supervised adversarial training $\eta$ in Eq.\ref{eq:FGSM} is set to 0.04.
The hyper-parameter that determines the number of sampled antibodies, denoted as $k$ in Algorithm~\ref{alg:antibody} is set to a value of 10 during the training.
The impact of $k$ will be discussed in Section~\ref{sec:Experiments}.


Stochastic gradient descent (SGD) with momentum is adopted for training with a batch size of 4.
The learning rate is set to 0.01, the momentum is set to 0.9.
%
The face recognition model $F(\cdot)$ in Eq.~\ref{eq:affinity} and Eq.~\ref{eq:loss} employs ArcFace (ResNet-50)~\cite{Deng2018ArcFace}.
%
Prior to conducting self-supervised adversarial training, a warm-up training phase of 50 thousand steps is performed.
During this phase, no noise was added to the input images, and the model is solely tasked with completing the facial reconstruction task.
Following the warm-up training, the self-supervised adversarial training is carried out for an additional 250 thousand steps.


\section{Experiments}
\label{sec:Experiments}

In this section, we evaluate the proposed approach through a series of experiments.
%
Firstly, we evaluated the defensive performance of the proposed method by employing the general adversarial attack methods and compared it with the state-of-the-art adversarial defense methods.
%
Following that, we evaluated the defensive performance of the proposed method using adversarial attack techniques tailored for the realm of facial recognition tasks.
%
To conduct a comprehensive examination of the proposed approach, we further employed adaptive attack strategies to rigorously test its efficacy.
%
Moreover, in order to delve deeper into the proposed approach, we conducted an extensive analysis of the generated antibodies by the model.
%
Finally, through ablation studies, we validated the effectiveness of the proposed model as well as the self-supervised adversarial training.



\subsection{Evaluation under General Attacking Methods}
\label{sec:ExpGeneral}

This subsection employs general adversarial attack methods to evaluate the proposed defense method.
We utilize the Equal Error Rate (EER) as the evaluation metric for recognition performance:
\begin{equation}    
    FRR = \frac{\sum \mathbbm{1}(S_p > T)}{N_p}
\end{equation}
\begin{equation}
    FAR = \frac{\sum \mathbbm{1}(S_n > T)}{N_n}
\end{equation}
where FRR is the false rejection rate, FAR is the false acceptance rate, $T$ is the similarity threshold, $S_p$ is the similarities of positive pairs, $N_p$ is the number of positive pairs, $S_n$ is the similarities of negative pairs, $N_n$ is the number of negative pairs.
EER refers to the FAR (or FRR) when the threshold $T$ is set in such a way that FAR equals FRR.
EER is a concise performance index that serves as a comprehensive evaluation of the discriminative ability of a face recognition model.


Gradient-based adversarial attacks are the most prevalent white-box strategies employed.
In the context of white-box attacks, the target model remains fully visible to the attack methods, rendering it an arduous test for defense methods.
We have chosen three gradient-based white-box attack methods for testing:
FGSM~\cite{Goodfellow2014Explaining} is a classic one-step adversarial attack approach;
DeepFool~\cite{Moosavi2016DeepFool} utilizes gradient signals in an iterative manner for adversarial attacking;
PGD~\cite{Madry2017Towards} is the most powerful first-order adversarial attack method. 
The magnitude of the adversarial noises is quantified by the ratio between the scale of the noise and the scale of the clean image: $I(\zeta) = ||\zeta|| / ||x||$, where $\zeta$ is the adversarial noises.
$I(\zeta)$ is set to 0.04 in the experiments of this subsection.

The experiments are conducted using two datasets: Labeled Faces in the Wild (LFW)~\cite{huang2008labeled} and MegaFace~\cite{Kemelmacher2016The}.
For the LFW dataset, we adhere to the official benchmark protocol\footnote{http://vis-www.cs.umass.edu/lfw/pairs.txt},  which involves selecting 3,000 positive pairs and 3,000 negative pairs of images for face verification.
Regarding MegaFace, we choose 80 identities from the subset \emph{facescrub} that have more than 50 images per subject.
From each identity, we randomly select 10 images for testing. Consequently, there are 7,200 positive pairs and 632,000 negative pairs available.
During testing, adversarial noises are introduced to one image within each pair, while the other image remains unchanged.
%
%



We select a set of representative adversarial defense methods for comparison in this subsection.
The first group of methods focuses on pixel space denoising, aiming to remove adversarial noises or restore clean images in the pixel space, including Quilting~\cite{moosavi2018divide}, TVM~\cite{guo2017countering}, PixelDefend~\cite{song2017pixeldefend}, MagNet~\cite{meng2017magnet}, and PIN~\cite{Ren2022Perturbation}.
The second group of methods focuses on enhancing the robustness of the recognition model.%
Among them, MTER~\cite{zhong2019adversarial} solely relies on the adversarial training strategy.
Additionally, HGD~\cite{liao2018defense} and Xie et al.~\cite{xie2019feature} combine specially designed model architectures with adversarial training to enhance the robustness of the recognition model.
The official implementation of PIN, HGD, and MTER is used in our experiments.
PixelDefend, MagNet, and Xie et al. are trained from scratch on the same dataset with the proposed model for a fair comparison.
%
For the denoising-based methods, ArcFace (ResNet-50)~\cite{Deng2018ArcFace} is uniformly employed as the recognition model for implementation.

\begin{table*}[t]
\begin{center}
\caption{Defensive performances on LFW (EER). Lower EER is preferable. The proposed method excels in all three adversarial attacking scenarios while simultaneously delivering comparable performance on clean facial images.}
\label{tab:ICLFW}
{
\begin{tabular}{c|c|c|c|c}
\hline
\hline
\bf{Defense Method}  & \bf{Clean}$\downarrow$ & \bf{FGSM}$\downarrow$ & \bf{DeepFool}$\downarrow$ & \bf{PGD}$\downarrow$ \\
\hline
 No Defense  & 0.44\% & 41.97\% &89.49\% & 99.71\%  \\
\hline
 Quilting~\cite{moosavi2018divide}  & 8.77\% & 9.04\% &25.10\% & 45.79\%  \\

  TVM~\cite{guo2017countering}  & 2.95\% & 19.94\% &73.21\% & 96.62\%  \\

  PixelDefend~\cite{song2017pixeldefend}  & 2.05\% & 18.09\% &70.31\% & 97.70\%  \\

  MagNet~\cite{meng2017magnet} & 1.51\% & 7.86\% &14.48\% &46.04\%  \\

  PIN~\cite{Ren2022Perturbation}  & 1.95\% & 6.15\% &7.86\% & 29.77\%  \\
\hline
 HGD~\cite{liao2018defense} & 1.08\% & 17.35\% & 20.48\% &49.69\%  \\

 Xie et al.~\cite{xie2019feature} & \bf{0.93}\% & 20.33\% &28.87\% & 31.29\%  \\
   MTER~\cite{zhong2019adversarial}  & 2.62\% & 10.03\% &24.89\% & 61.06\%  \\
\hline
 Ours  & 1.01\% & \bf{4.46}\% &\bf{5.01}\% & \bf{14.19}\%  \\

\hline
\hline
\end{tabular}}
\end{center}
\end{table*}


\begin{table*}[t]
\begin{center}
\caption{Defensive performances on MegaFace (EER). Lower EER is preferable. The proposed approach demonstrates superior performance under most adversarial attacking, particularly showing a significant advantage against challenging PGD attacks.}
\label{tab:ICMega}
{
\begin{tabular}{c|c|c|c|c}
\hline
\hline
\bf{Defense Method}      & \bf{Clean}$\downarrow$ & \bf{FGSM}$\downarrow$ & \bf{DeepFool}$\downarrow$ & \bf{PGD}$\downarrow$ \\
\hline
No Defense                      & 1.31\%                          &  50.87\%  &  95.41\%    &  99.09\%   \\
\hline
 Quilting~\cite{moosavi2018divide}         & 14.23\%                    & 18.08\%     &  20.54\%   & 47.13\%  \\

  TVM~\cite{guo2017countering}            & 3.66\%                      & 21.59\%     & 76.24\%    & 95.12\%  \\

  PixelDefend~\cite{song2017pixeldefend}    & 2.41\%                        & 25.10\%     &  77.75\%    & 94.03\%  \\

  MagNet~\cite{meng2017magnet}             & 2.51\%                    & 10.79\%       & 24.51\%     &  47.11\%  \\

PIN~\cite{Ren2022Perturbation}             & 3.78\%    & 7.82\%     &    9.20\%    &    33.37\%  \\
\hline
 HGD~\cite{liao2018defense}               & 2.44\%                            & 16.15\%       & 29.09\%     &  52.27\%  \\

 Xie et al.~\cite{xie2019feature}          & \bf{1.89}\%                     & 16.21\%       &  34.27\%     & 48.76\%  \\

 MTER~\cite{zhong2019adversarial}          & 3.08\%                    & 11.50\%       &  33.44\%     & 66.95\%  \\
\hline
  Ours                                  & 2.23\%    & \bf{7.27}\%     &    \bf{9.17}\%    &    \bf{18.63}\%  \\

\hline
\hline

\end{tabular}}
\end{center}
\end{table*}


The experimental results on the LFW dataset are presented in Table~\ref{tab:ICLFW}, while the results on the MegaFace dataset are shown in Table~\ref{tab:ICMega}.
The proposed adversarial defense method demonstrates superior defense performance on both datasets in most cases.
From the experimental results, it can be observed that denoise-based methods often face a trade-off between the performance on clean samples and adversarial defense performance, making it difficult to strike a balance between the two sides.
This is because these methods typically attempt to employ the same denoising approach to counter all adversarial noises, making it challenging to cope with the diversity and complexity of adversarial noises in face recognition.
Although the adversarial training based methods achieve satisfactory performance under FGSM attacking, its performance fluctuates significantly under different attacking methods, indicating a lack of generalization to different types of adversarial noises.
In contrast, the proposed method is capable of achieving superior defense performance while only sacrificing a slight decrease in recognition performance on clean face images comparing to Xie et al.~\cite{xie2019feature}.
This is attributed to the fact that adversarial training methods (including Xie et al.~\cite{xie2019feature}) retrain the facial feature extractor, thereby maintaining its performance on clean images, whereas the proposed method based on noise removal operate under the premise of a fixed facial feature extractor.
%

%


\subsection{Evaluation under Attacking Methods Tailored for Face Recognition}
%
Due to the importance and uniqueness of facial recognition tasks, researchers have proposed adversarial attack methods specifically targeting facial recognition.
In the real-world scenarios, black-box attacks are more common than white-box attacks since the former do not require access to the recognition model's information, making their execution plainer and more straightforward.
To ascertain the effectiveness of the proposed method against black-box attacks, we experimented with various forms of black-box attacks tailored for face recognition in this subsection.
%
%
Compared to general adversarial attack methods, these methods pose a greater threat to the practical application of facial recognition systems.
%
%


The adversarial attack methods employed include the following three:
\emph{DFANet}~\cite{zhong2020toward}: DFANet is a transfer-based black-box attack method specifically designed for facial recognition models.
It adopts gradient-based attack techniques for facial recognition tasks, making it a powerful transfer-based attacking method for face recognition.
\emph{Evolutionary}~\cite{Dong2019Efficient}: Evolutionary is a black-box attack method targeting facial recognition systems.
It directly exploits the decision outcomes of the facial recognition system, without requiring access to the gradient information of the recognition model.
%
\emph{Sticker attacking}~\cite{Komkov2013AdvHat, Yang_2023_CVPR}: Sticker attacking is a category of black-box attack methods that operate in the physical domain.
By making subtle modifications to the physical appearance of the target object, it produces powerful attacks and can be easily carried out by an attacker without any prior knowledge about the recognition model.
These methods significantly reduce the cost of adversarial attacks, posing a substantial threat to facial recognition systems.
%
The comparative methods employed in this subsection remain consistent with the ones used in Section~\ref{sec:ExpGeneral}.
Similarly, ArcFace is employed as the recognition model for conducting experiments on denoising-based methods as well.


\paragraph{DFANet:}
The experiments under the DFANet attacking are conducted using the TALFW dataset~\cite{TALFW}.
The adversarial samples of TALFW are crafted by utilizing DFANet, with the source images being obtained from LFW.
TALFW serves as an official test benchmark offered by DFANet.
%
The evaluation on TALFW still utilizes EER as the performance index for assessment.

\begin{table*}[t]
\begin{center}
\caption{The experimental results on TALFW. The proposed method demonstrates the most outstanding defensive performance.}
\label{tab:TALFW}
\setlength{\tabcolsep}{10mm}
{
\begin{tabular}{c|c}
\hline
\hline
\bf{Defense Method}  & \bf{EER}$\downarrow$  \\
\hline
 No Defense  & 39.71\%   \\
\hline
 Quilting~\cite{moosavi2018divide}  & 24.33\%   \\

  TVM~\cite{guo2017countering}  & 30.90\%   \\

  PixelDefend~\cite{song2017pixeldefend}  & 33.22\%   \\

  MagNet~\cite{meng2017magnet} & 22.50\%    \\

  PIN~\cite{Ren2022Perturbation}  & 21.95\%   \\
\hline
 HGD~\cite{liao2018defense} & 39.23\%   \\

 Xie et al.~\cite{xie2019feature} & 40.33\%   \\
   MTER~\cite{zhong2019adversarial}  & 39.16\%   \\
\hline
 Ours  & \bf{20.43}\%   \\

\hline
\hline
\end{tabular}}
\end{center}
\end{table*}

The testing results on TALFW are presented in Table~\ref{tab:TALFW}.
The proposed method demonstrates superior defensive performance compared to the comparative methods.
Through experimentation, it can be observed that methods based on adversarial training perform almost on par with the performance without any defense mechanisms, exhibiting a significant gap when compared to methods based on denoising.
%
%
Adversarial training based methods extensively incorporate adversarial samples into the training set and continually seek out vulnerabilities in the recognition model.
However, since adversarial training is a dynamic process with new data constantly emerging, the model often focuses only on the latest training data during the training process, which can lead to a situation where fixing one problem leads to another.
This results in the performance of the model being nearly identical to that of models without any defense mechanisms when facing attacking of DFANet.
On the other hand, the proposed memory module possesses the capacity to preserve adversarial noise patterns, thereby mitigating the challenges associated with adversarial training throughout the self-supervised adversarial training procedure.


\paragraph{Evolutionary:}
When employing the Evolutionary method to adversarial attacks, the attacking process revolves around continuously optimizing the adversarial samples under the constraint of achieving successful attacks.
The objective is to gradually minimize the disparity between the adversarial samples and the clean facial images.
Hence, we utilize the defense performance metric proposed officially within the realm of Evolutionary methods, namely the mean squared error (MSE) between the adversarial samples and the clean facial images under the same number of optimization steps.
The larger this error, the more challenging the attack becomes, signifying a stronger defense performance.
%
There are two experimental configurations: dodging and impersonation.
Dodging refers to adversarial attacks attempting to recognize positive facial image pairs as negative ones, while impersonation is the opposite, attempting to recognize negative facial image pairs as positive ones.
The experimental testing dataset employed in this experiment remains consistent with those in Section~\ref{sec:ExpGeneral}.
%
Due to their high computational complexity, Quilting, TVM, and PixelDefend are difficult to implement for Evolutionary attacking experiments.
Consequently, we select the remaining comparative methods discussed in Section~\ref{sec:ExpGeneral} for this experimentation.

\begin{table*}[t]
\begin{center}
\caption{MSE on LFW under Evolutionary attacking. At the same step of the iteration, a higher average distortion indicates better performance of the defense method. The proposed method surpasses the listed comparative methods.}
\label{tab:FRBlackLFW}
{
\begin{tabular}{c|c|c|c|c}
\hline
\hline
\multicolumn{2}{c|}{\bf{Number of Attack Steps}}  & \bf{1000} $\uparrow$ & \bf{5000}$\uparrow$ & \bf{10000}$\uparrow$ \\
\hline
\multirow{6}*{Dodging} & No Defense                   & 3.1e-3 & 2.5e-4 & 8.9e-5  \\
&MagNet~\cite{meng2017magnet}                            &  1.0e-2 & 7.7e-3 & 5.8e-3  \\

& PIN~\cite{Ren2022Perturbation}  & 7.5e-2 & 5.5e-2 & 3.1e-2    \\

& HGD~\cite{liao2018defense}                                & 8.7e-3 & 6.3e-3 & 3.9e-3\\
& Xie et al.~\cite{xie2019feature}                                & 7.4e-3 & 5.8e-3 & 2.7e-3 \\
&MTER~\cite{zhong2019adversarial}    & 9.9e-3 & 6.5e-3 & 4.9e-3  \\
&Ours                       & \bf{1.1e-1} & \bf{7.5e-2} & \bf{4.7e-2}  \\
\hline
\hline
\multirow{6}*{Impersonation} & No Defense             &  2.4e-3 & 2.4e-4 & 5.9e-5  \\
&MagNet~\cite{meng2017magnet}                            & 8.3e-3 & 5.7e-3 & 3.1e-3  \\
& PIN~\cite{Ren2022Perturbation}  & 4.6e-2 & 3.6e-2 & 2.9e-2    \\
& HGD~\cite{liao2018defense}                                & 1.1e-3 & 7.2e-4 & 4.8e-4 \\
& Xie et al.~\cite{xie2019feature}                 & 9.7e-4 & 6.6e-4 & 4.1e-4 \\
&MTER~\cite{zhong2019adversarial}    & 1.7e-3 & 9.3e-4 & 6.1e-4  \\
&Ours         &  \bf{1.3e-1} & \bf{9.2e-2} & \bf{7.9e-2}  \\
\hline
\hline
\end{tabular}}
\end{center}
\end{table*}


\begin{table*}[h]
\begin{center}
\caption{MSE on MegaFace under Evolutionary attacking. It particularly demonstrates the advantages of the proposed method when the number of iterations increases, as it provides a better reflection of the performance of defense methods.}
\label{tab:FRBlackMega}
{
\begin{tabular}{c|c|c|c|c}
\hline
\hline
\multicolumn{2}{c|}{\bf{Number of Attack Steps}}  & \bf{1000}$\uparrow$ & \bf{5000}$\uparrow$ & \bf{10000}$\uparrow$ \\
\hline
\multirow{6}*{Dodging} & No Defense                    & 3.5e-3 &  8.5e-4&   9.7e-5\\

&MagNet~\cite{meng2017magnet}                      & \bf{9.8e-2} & 6.5e-3 & 2.2e-3  \\
& PIN~\cite{Ren2022Perturbation}  & 9.2e-2 & \bf{8.4e-2} & 6.7e-2    \\
& HGD~\cite{liao2018defense}                                & 8.8e-2 & 6.0e-3 & 1.9e-3 \\
& Xie et al.~\cite{xie2019feature}                                & 7.2e-3 & 5.5e-3 & 2.3e-3 \\
&MTER~\cite{zhong2019adversarial} & 9.7e-3 & 6.6e-3 & 5.1e-3  \\
&Ours                              & 9.5e-2 & 8.3e-2 & \bf{6.9e-2}  \\
\hline
\hline
\multirow{6}*{Impersonation} & No Defense           & 2.4e-3 &  1.1e-4& 5.5e-5  \\
&MagNet~\cite{meng2017magnet}                                        & 7.9e-3 & 2.1e-3 & 1.4e-3\\
& PIN~\cite{Ren2022Perturbation}  & \bf{7.7e-2} & 6.8e-2 & 6.2e-2    \\
& HGD~\cite{liao2018defense}                                & 7.2e-3 & 1.7e-3 & 8.4e-4 \\
& Xie et al.~\cite{xie2019feature}                                & 6.5e-3 & 3.7e-3 & 1.8e-3 \\
&MTER~\cite{zhong2019adversarial} & 7.9e-3 & 4.6e-3 & 2.9e-3  \\
&Ours           & 7.6e-2 & \bf{7.1e-2} & \bf{6.6e-2}  \\
\hline
\hline
\end{tabular}}
\end{center}
\end{table*}


The experimental results are shown in Table~\ref{tab:FRBlackLFW} and Table~\ref{tab:FRBlackMega}.
The experimental results demonstrate that the proposed approach outperforms other methods in most cases across both datasets.
As Evolutionary attack relies on continuously exploring the decision boundaries of the target recognition model to optimize adversarial noises, the performance metrics at larger numbers of attack iterations better reflect the effectiveness of defense methods.
The advantage of the proposed approach becomes more prominent with a larger number of attack steps.
%
%
This is attributed to the fact that the proposed adversarial defense methodology exhibits the capacity to analyze and retain crucial information from each facial sample, leading to a more robust decision boundary.
As a corollary, the task of mounting the Evolutionary attack becomes noticeably more formidable.


\paragraph{Sticker Attacking:}
%
Sticker attacking is a category of physical domain adversarial attack methods, where adversarial stickers are applied to specific areas of faces to introduce adversarial noises during the image acquisition process.
Adversarial stickers can be directly applied to the face~\cite{Yang_2023_CVPR} or attached to accessories~\cite{komkov2021advhat}, and can even be achieved through makeup techniques~\cite{Zhu2019Generating}.
These methods do not require intervention in the data processing pipeline of the face recognition system, making them cost-effective and posing a threat to the actual deployment of face recognition systems.

In practical scenarios, both dodging and impersonating types of sticker attacks are prevalent.
But these two forms of attack typically target different modules of face recognition systems.
Dodging attacks usually deceive the face detection module to prevent the attacker from being detected, while impersonating attacks, conducted after a face has been successfully detected, usually confuse the facial feature extraction component to misidentify the attacker as another individual.
Actually, simply obscuring key regions of the face (e.g., by wearing masks or hats) suffices to achieve the desired result of dodging attacks.
If faces can not be successfully detected, adversarial defenses for facial feature extraction are not needed.
Given that this work focuses on adversarial defenses in facial feature extraction, we have specifically conducted experiments on impersonating attacks.

\begin{figure}[t]
\begin{center}
\includegraphics[width=0.8\linewidth]{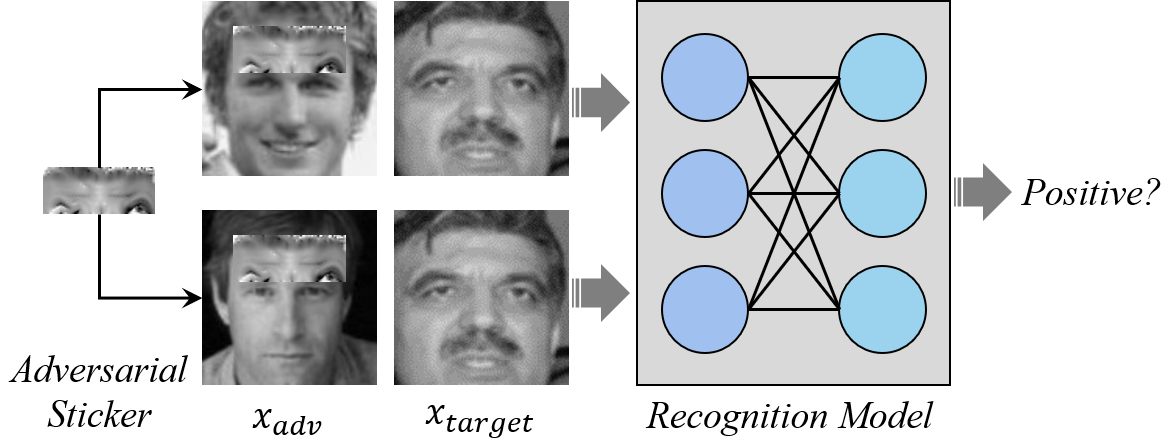}
\end{center}
   \caption{The sticker attacking employs the adversarial sticker strategically overlaid on specific regions of facial images to induce mistakes in recognition models.}
\label{fig:sticker}
\end{figure}

In our experiments, we employ AdvHat~\cite{komkov2021advhat} as the adversarial attack method.
We simulated the generation process of adversarial stickers and then applied the generated stickers to the facial images for the attack, as shown in Fig.~\ref{fig:sticker}.
Unlike other adversarial attack methods, the sticker attack does not impose a restriction on the intensity of adversarial noises.
%


A total of 1000 facial images from different individuals are randomly selected from the LFW for the test.
The generated adversarial sticker is applied on the foreheads of these test facial images.
The objective of the adversarial attack is to manipulate the facial recognition model into identifying all 1000 test faces as the target identity.
The size of the adversarial sticker was set to $20 \times 72$.
The decision similarity threshold for facial recognition was deliberately set at 0.2 to ensure that the recognition model (ArcFace) correctly identifies the majority of positive pairs while maintaining an acceptable false negative rate (TAR=99.89\%, FRR=2.09\%) on clean facial images.
This stringent configuration serves as a rigorous benchmark for evaluating defense methods.
%
The accuracy of recognition serves as the performance index in the experimentation.
When the similarity between a test facial image with the added adversarial sticker and the target facial image is below 0.2, it is considered a correct identification. 
Conversely, if the similarity exceeds 0.2, it is considered an incorrect identification.
A higher accuracy indicates better performance of the defense method.

\begin{table*}[t]
\begin{center}
\caption{The accuracy of recognition under sticker attacking. The proposed method surpassed all comparative methods, yielding the best results.}
\label{tab:sticker}
\setlength{\tabcolsep}{10mm}
{
\begin{tabular}{c|c}
\hline
\hline
\bf{Defense Method} & \bf{Accuracy} $\uparrow$  \\
\hline
No Defense & 25.6\%    \\
\hline
Quilting~\cite{moosavi2018divide} & 92.2\%  \\

TVM~\cite{guo2017countering} & 86.0\%   \\

PixelDefend~\cite{song2017pixeldefend} & 92.7\%    \\

MagNet~\cite{meng2017magnet} & 95.9\%    \\

PIN~\cite{Ren2022Perturbation}  & 97.6\%    \\
\hline

HGD~\cite{liao2018defense}   & 89.1\%  \\

Xie et al.~\cite{xie2019feature}     & 90.7\%  \\

MTER~\cite{zhong2019adversarial} & 93.4\%    \\
\hline 

Ours & \bf{98.4}\% \\
\hline
\hline
\end{tabular}}
\end{center}
\end{table*}

The recognition accuracy under sticker attacking is shown in Table~\ref{tab:sticker}.
Experimental results demonstrate that the proposed method achieved the best recognition outcomes.
The adversarial noises in the physical domain differ significantly from that in the digital domain, as their intensity is no longer constrained, posing a great challenge for noise removal.
Due to the significant alterations caused by adversarial stickers, the resulting facial images deviate notably from the distribution of clean facial images.
Therefore, the key to denoising in such cases lies in effectively modeling and utilizing the distribution of clean facial images.
The proposed method achieves this through the designed antibodies.
%
Moreover, the proposed adversarial defense approach offers an ability to provide customized defensive measures for each individual sample, making it arduous for adversarial stickers to have a universal impact across diverse facial samples.
As a result, the practical applicability of sticker attacks in real-world scenarios is significantly diminished.


\subsection{Evaluation under Adaptive Attacking}

For real-world deployment of face recognition systems, if attackers become aware of adversarial defense mechanisms, they are likely to face adaptive attacking.
An adaptive attack poses a formidable challenge.
In this type of attack, attackers not only target the recognition model but also optimize their strategies to circumvent the defense mechanisms.
These attacks rigorously test the efficacy of defense methods, as attackers have the capability to adjust their tactics based on the employed defenses, leaving the defense mechanisms unable to adapt in response.


In this subsection, we present an adaptive attack against the proposed adversarial defense method to evaluate its defensive capabilities.
Given that the proposed defense method is built upon antibodies, the composition of antibodies serves as the foundation of the entire method.
Therefore, in the adaptive attack, all possible antibodies involved in the proposed method are considered within the attack scope.
Specifically, during the optimization of adversarial noises, the input facial images undergo preprocessing with a specific antibody $a^*$, which consists of the top 1500 eigenvectors with the largest eigenvalues:
\begin{equation}
    a^* = \{ e_1, e_2, ..., e_{1500} \}
\end{equation}
This means that all possible antibodies are subsets of $a^*$, which means the optimization direction will encourage the adversarial noises to avoid being removed by any possible antibodies.
This poses a rigorous test for the proposed method.
%
We utilized the three adversarial attack methods outlined in Section~\ref{sec:ExpGeneral} for the optimization of adversarial noises in the adaptive attack.
The intensity of the adversarial noises $I(\zeta)$ is also set as 0.04, and the experiment is conducted on LFW.

\begin{table*}[t]
\begin{center}
\caption{The experimental results of the adaptive attacking (EER).}
\label{tab:Adaptive}
{
\begin{tabular}{c|c|c|c}
\hline
\hline
\bf{Attack Method}   & \bf{FGSM}$\downarrow$ & \bf{DeepFool}$\downarrow$ & \bf{PGD}$\downarrow$ \\
\hline
 No Defense  & 16.39\% & 33.55\% & 54.86\%   \\
\hline
 Ours        & 10.76\% & 24.92\% &  38.68\%   \\

\hline
\hline
\end{tabular}}
\end{center}
\end{table*}


The experimental results are presented in Table~\ref{tab:Adaptive}.
Comparing with Table~\ref{tab:ICLFW}, the experimental results reveal a significant decline in recognition performance of the proposed defense method when subjected to adversarial noise generated by the adaptive attacking.
This can be attributed to the targeted nature of the adaptive attacking, which diminishes the effectiveness of the antibodies generated by the defense method.
However, even under the adaptive attacking, the proposed method still outperforms most of the comparative methods listed in Table~\ref{tab:ICLFW}.
Furthermore, when contrasting with the first row of Table~\ref{tab:ICLFW}, it is evident that the adaptive attacking have a significantly reduced impact on the model without any adversarial defense measures.
This indicates the substantial cost incurred by adversarial noises in bypassing the proposed defense method, resulting in a greatly weakened attack effect in the absence of any defense measures.
This further highlights the superior defensive performance of the proposed method.


\subsection{Quantitative Analysis of Antibodies}

In order to conduct a more comprehensive analysis of the proposed defense method, this section focuses on the quantitative analysis of the antibodies generated by the proposed method, unveiling the evolutionary process of antibodies during model training.
This analysis primarily encompasses three aspects of antibodies: their sparsity, mutation probability, and specificity.



\paragraph{Sparsity of Antibodies:}
%
The sparsity of antibodies is determined by the number of eigenvectors they contain, and a smaller quantity of eigenvectors indicates a higher level of sparsity in the antibodies.
The number of eigenvectors present in antibodies represents their selection of features for noise removal and facial sample reconstruction.
A greater number of eigenvectors suggests that the antibodies are biased towards reconstructing more intricate details of the input samples, while a smaller number indicates their inclination towards extracting a smaller subset of essential features.
To quantitatively measure the sparsity of antibodies, we employ the number of eigenvectors contained within them as a metric to evaluate their sparsity.

\begin{figure}[t]
\begin{center}
\includegraphics[width=0.9\linewidth]{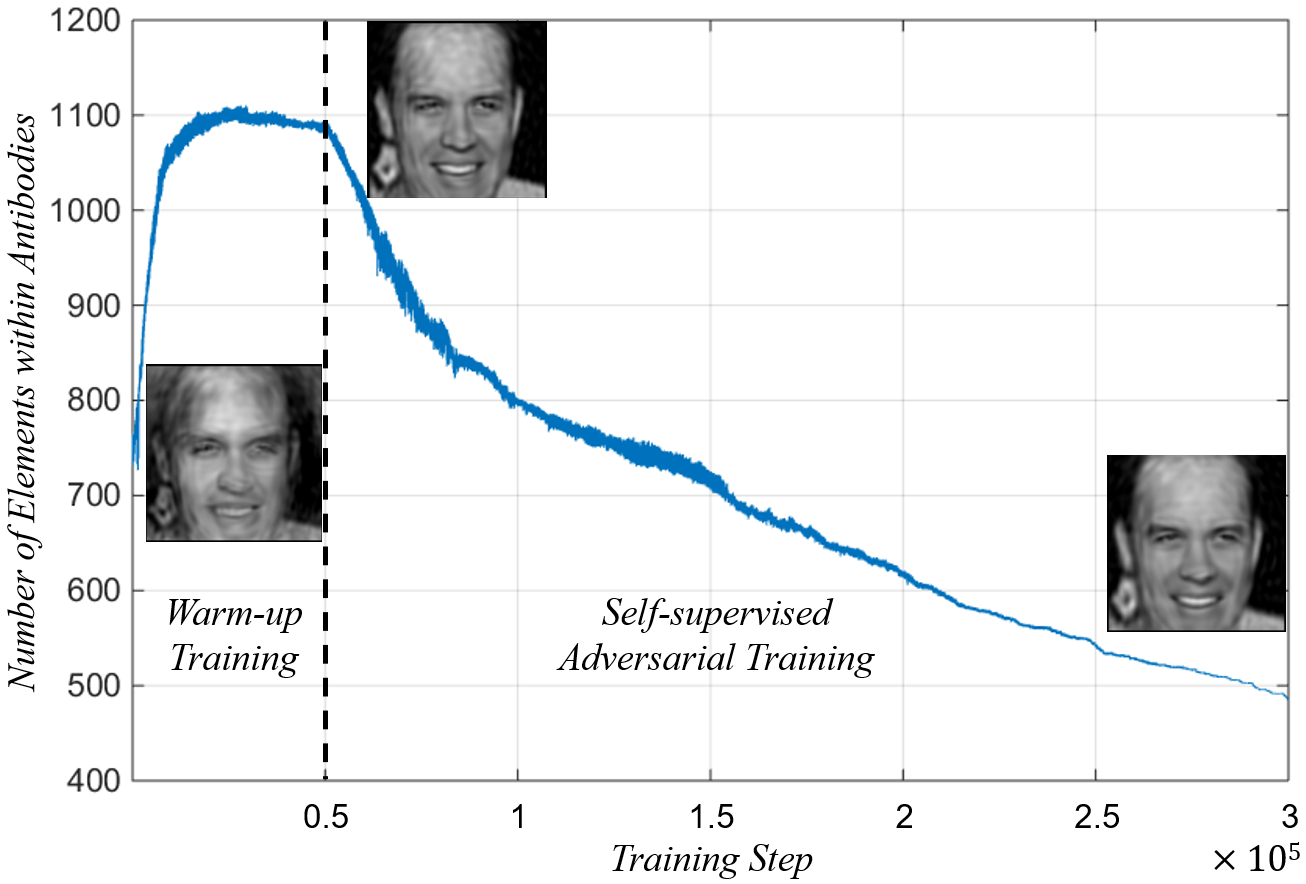}
\end{center}
   \caption{The number of eigenvectors contained within antibodies during the training process. During the training process, the number of eigenvectors present in the antibodies initially rises and then declines. This progression enables the antibodies to first enhance their reconstruction capabilities, followed by a selective refinement of eigenvectors to bolster their denoising prowess. As a result, the antibodies achieve the remarkable ability to effectively eliminate adversarial noise while preserving vital facial features.}
\label{fig:sparsity}
\end{figure}

\begin{figure}[h]
\begin{center}
\includegraphics[width=0.5\linewidth]{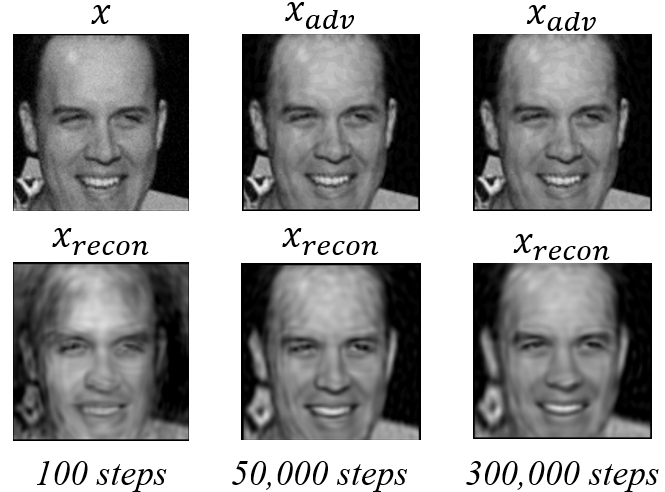}
\end{center}
   \caption{Examples of the recovery performance of antibodies on input face images at different stages of training.}
\label{fig:recon_step}
\end{figure}

The number of eigenvectors contained within antibodies during the entire training process is depicted Fig.~\ref{fig:sparsity} (average number of antibodies per mini-batch).
%
Several key observations can be summarized from Fig.~\ref{fig:sparsity}:
%
\begin{itemize}
    \item During the initial warm-up training phase, the number of eigenvectors in antibodies rapidly increases.
    This is because, at this stage, the model focuses solely on facial image reconstruction without incorporating self-supervised adversarial training.
    By increasing the number of eigenvectors in antibodies, the model can reduce reconstruction errors.
    Therefore, during antibody optimization, the model quickly increases the number of eigenvectors, thereby reducing antibody sparsity and improving reconstruction results.

    \item In the warm-up training phase, the number of eigenvectors in antibodies reaches around 1100 and fluctuates around this value.
    This occurs due to the regularization term in Eq.\ref{eq:affinity}, which penalizes the number of elements in antibodies.
    The benefits of adopting more eigenvectors are suppressed by the penalty imposed by the regularization term in Eq.\ref{eq:affinity}.

    \item Once self-supervised adversarial training is introduced, the number of eigenvectors in antibodies gradually decreases, indicating an increase in antibody sparsity.
    This is because, in self-supervised adversarial training, the model no longer focuses solely on reconstructing input samples but tracks its target at removing adversarial noise by refining the eigenvectors.

    \item The Fig.\ref{fig:recon_step} illustrates the recovery effects of antibodies at different training stages.
    When training concludes, the number of eigenvectors in generated antibodies has decreased to around 500.
    Although this number is lower compared to the initial 750, the recovery effects have significantly improved compared to the initial phase of training.
    Compared to the end of the warm-up training phase, the antibodies at the final stage retain crucial facial information and effectively remove adversarial noise by filtering out some eigenvectors.
\end{itemize}

Through the aforementioned analysis, it can be observed that the number of eigenvectors contained within the antibodies initially increases and then decreases during the entire training process.
However, during this process, the defense model does not stand still but goes through a process of first improving its reconstruction ability and then refining the denoising ability by eliminating crude components from the eigenvectors.
Eventually, this results in an effective removal of adversarial noise while preserving the key features of the face.

\paragraph{Mutation Probability:}
%
During the optimization process of the proposed defense model, the antibody mutation plays a crucial role.
Adequate antibody mutation during training not only enhances antibody diversity but also effectively prevents the model from getting trapped in local optima during optimization.
However, if the degree of antibody mutation is excessively high, it may lead to difficulties in the convergence of the defense model.
In order to further investigate the optimization process of the proposed defense method, we analyze the probability of antibody mutation.


The antibody mutation refers to the changes that occur in the eigenvectors comprising the antibodies.
In order to quantitatively measure the probability of antibody mutation, we calculate the average probability of reversing the selection of antibodies for each eigenvector (whether to include it or not) as a metric of evaluation:
\begin{equation}
    P_{mutation} = \frac{1}{k}\sum^k_i (0.5-|(f_e(i)-0.5)|)
\end{equation}
where $k$ is the dimension of $f_e$, $|\cdot|$ refers to absolute value.
%
The closer $f_e(i)$ is to 0.5, the more likely it is for the selection of the $i$-th eigenvector to undergo mutation.

\begin{figure}[t]
\begin{center}
\includegraphics[width=0.9\linewidth]{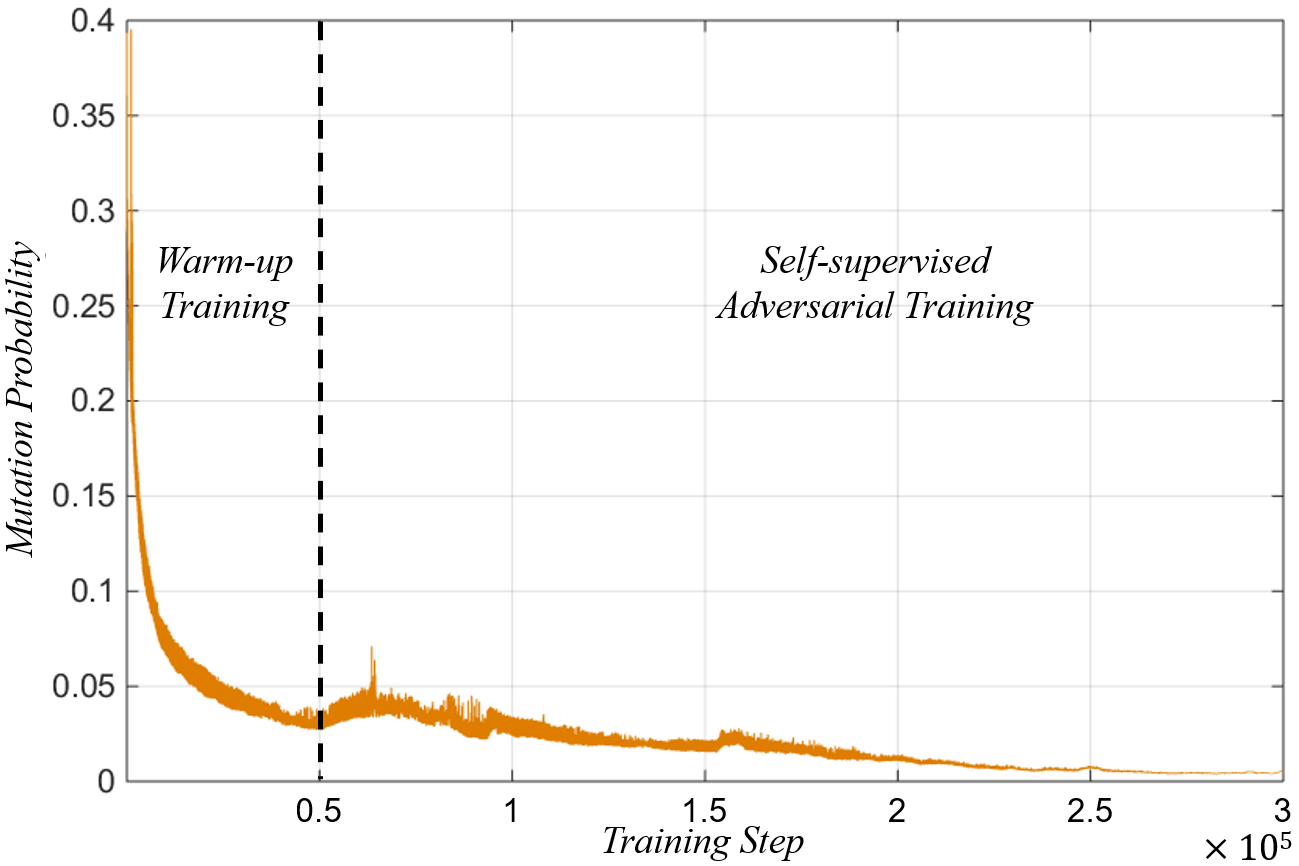}
\end{center}
   \caption{The mutation probability of antibodies during the training process. The proposed defense approach demonstrates the ability to adaptively adjust the mutation probability of antibodies during the training process, aligning them with the evolving training objective. When the training objective undergoes changes, the model responds by increasing the mutation probability, thereby enhancing the overall flexibility of the defense model. Conversely, as the training objective stabilizes,  the model gradually reduces the mutation probability of antibodies, allowing for a progressive convergence of the model.}
\label{fig:mutation}
\end{figure}


The mutation probability of antibodies during the entire training process is shown in Fig.~\ref{fig:mutation} (the mean of each mini-batch).
From Fig.~\ref{fig:mutation}, we can draw several conclusions:
\begin{itemize}
    \item During the initial state, the probability of mutation for antibodies is high, but during the warm-up training phase, this probability decreases rapidly.
    This is due to the model quickly honing in on the critical eigenvectors for face image reconstruction, resulting in a rapid decrease in the mutation probability of these eigenvectors.

    \item Upon entering the self-supervised adversarial training phase, the mutation probability of antibodies slightly increases.
    This is due to a change in the training objective of the model, and the model adapts by increasing the mutation probability of antibodies to better fit this new objective.

    \item The mutation probability of antibodies eventually converges and approaches 0, indicating that the model gradually finds the best antibody for each input sample after training, and no longer mutates.
\end{itemize}

Through the fluctuations in the mutation probability of antibodies, we can observe how the proposed defense method possesses the capability to adaptively modify the mutation probability to align with the evolving training objective.
As the training objective undergoes changes, the model augments the mutation probability, promoting the flexibility of the defense system and diversifying the antibodies.
Consequently, this enables a readaptation to the new objective at hand.
Conversely, when the training objective stabilizes, the model progressively diminishes the mutation probability of antibodies, fostering convergence.
Thus, the proposed method exhibits dynamic adaptability akin to that of an immune system.

\paragraph{Specificity of Antibody:}
%
Due to the particular nature of facial recognition tasks, the specificity and diversity of adversarial noises in facial recognition are remarkably pronounced.
Therefore, we aspire for the defense approach to furnish tailored noise removal strategies for each input facial image, thereby addressing this formidable challenge. 
In order to investigate whether the proposed defense method indeed confers specificity to the antibodies, we examine the specificity of antibodies during the training process.


The specificity of antibodies can be discerned through differences manifested between antibodies of different samples.
To quantitatively assess the specificity of antibodies, we propose a metric that measures the dissimilarity between two antibodies:
\begin{equation}
    J(a_i, a_j) = \# \{ e| (e\in a_i \wedge e \notin a_j) \vee (e\notin a_i \wedge e \in a_j) \}
\end{equation}
%
which means the number of eigenvectors that are uniquely present in either one of the antibodies.
Based on $J(a_i, a_j)$ it is possible to quantitatively measure the specificity of a set of antibodies:
\begin{equation}
    V(\{ a_1, a_2, ..., a_n \}) = \frac{1}{n(n-1)}\sum^n_i \sum^n_{j\neq i} J(a_i, a_j)
\end{equation}
%


\begin{figure}[t]
\begin{center}
\includegraphics[width=0.9\linewidth]{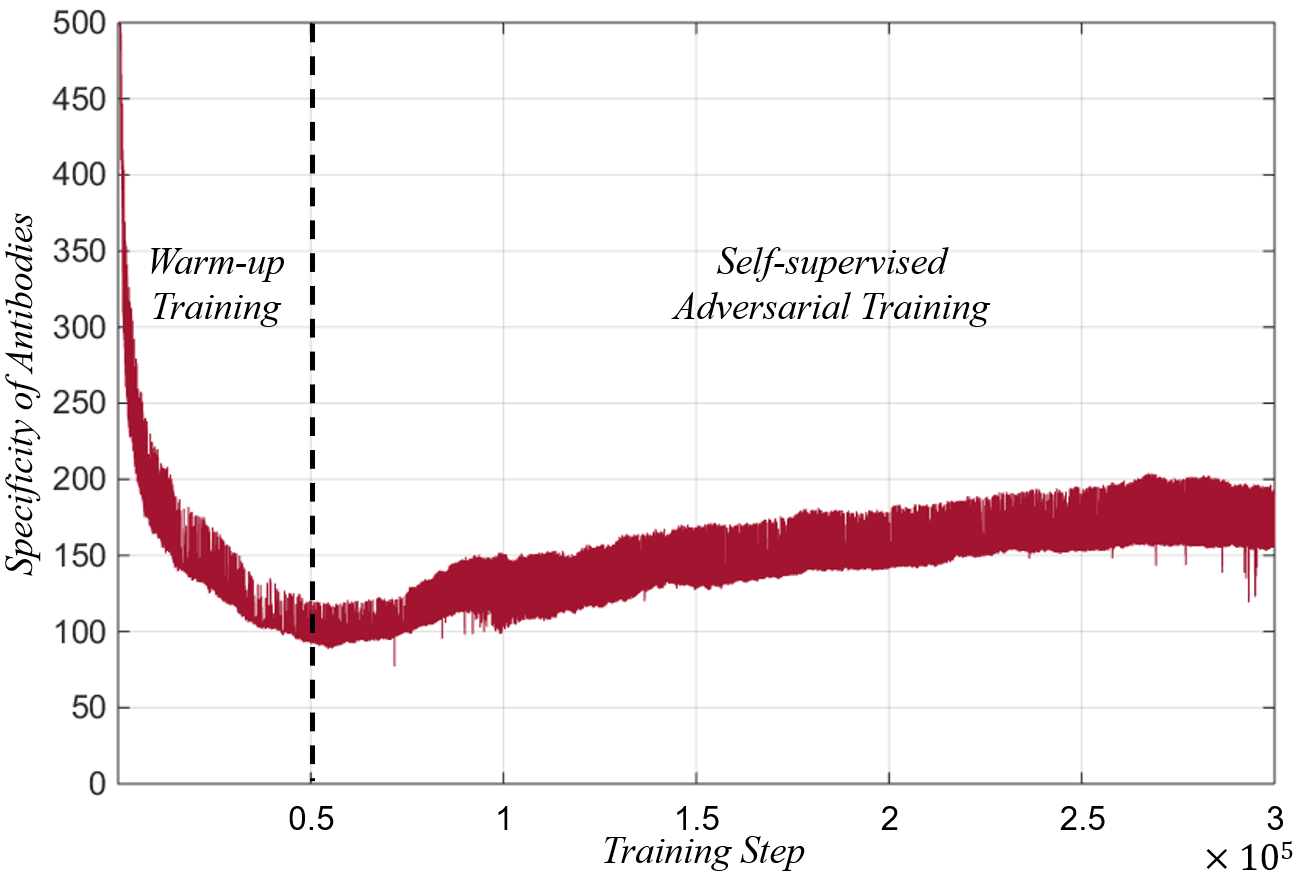}
\end{center}
   \caption{The specificity of antibodies during the training process. The specificity of antibodies undergoes a process of initial decline followed by a subsequent recovery during the training. The decrease in specificity during the warm-up training phase can be primarily attributed to the rapid reduction in antibody mutation probability. On the other hand, the gradual increase in antibody specificity observed during the self-supervised adversarial training is a result of the model progressively providing tailored noise removal strategies for each individual sample to effectively eliminate adversarial noise.}
\label{fig:specificity}
\end{figure}


The specificity of each mini-batch of antibodies throughout the entire training process is illustrated Fig.~\ref{fig:specificity}.
Several observations can be made from Fig.~\ref{fig:specificity}:
\begin{itemize}
    \item At the beginning of the training process, the specificity of the antibodies is initially high, due to the high probability of antibody mutation (as shown in the Fig.~\ref{fig:mutation}) and the instability of the antibodies.
    After the warm-up training, the specificity rapidly decreases to 100.
    The reason is that facial images contain shared characteristics and the goal of the warm-up training is to reconstruct facial images, rendering high antibody specificity unnecessary.

    \item Upon commencing self-supervised adversarial training, the specificity of the antibodies undergoes a gradual enhancement and eventually settles at approximately 200.
    This implies that, on average, there are 200 distinct eigenvectors between every pair of antibodies.
    It indicates that the defense model progressively commences providing noise removal manners tailored to each specific adversarial sample during the course of self-supervised adversarial training.

\end{itemize}
    
Although the specificity of the antibodies experienced a process of decline and then recovery during the training process, this does not mean that the defense model returned to the initial state.
The decline in specificity observed during the warm-up training phase primarily stems from the swift reduction in antibody mutation probability and the enhancement of antibody stability.
Conversely, the gradual augmentation of antibody specificity throughout the self-supervised adversarial training arises from the model's provision of tailored noise removal solutions for each individual sample, thereby eliminating adversarial noises.


\subsection{Ablation Study}

In this subsection, we first verify through experiments the impact of self-supervised adversarial training and the memory module on defense performance, then explore the applicability of the proposed defense method to different recognition models.
Finally, we conduct sensitivity analysis on key hyper-parameters in the model optimization process.

\paragraph{Ablation Study for Self-supervised Adversarial Training:}
%
Self-supervised adversarial training is a crucial component of the proposed defense method.
Its purpose is to enhance the noise removal ability of defense models by introducing adversarial noises during the training process.
To evaluate the effectiveness of this training strategy, we compared the model without self-supervised adversarial training to the original model.
We conducted experiments on LFW using three types of adversarial attacks: FGSM, DeepFool, and PGD.
%
The experimental protocols are consistent with Section~\ref{sec:ExpGeneral}.

\begin{table*}[t]
\begin{center}
\caption{The experimental results of ablation study for self-supervised adversarial training and memory module (EER). SSAT refers to self-supervised adversarial training. These results also demonstrate the significant improvement in the defense capability of the model by self-supervised adversarial training and memory module.}
\label{tab:abl}
{
\begin{tabular}{c|c|c|c|c}
\hline
\hline
\bf{Attack Method} & \bf{Clean}$\downarrow$ & \bf{FGSM}$\downarrow$ & \bf{DeepFool} $\downarrow$ & \bf{PGD}$\downarrow$ \\
\hline
 w/o SSAT    & 1.04\% & 11.99\% & 12.08\% & 40.60\%   \\
 \hline
 w/o Memory    & 1.06\% & 9.06\% & 9.75\% & 22.41\%   \\
\hline
 Ours        & \bf{1.01}\% & \bf{4.46}\% &\bf{5.01}\% & \bf{14.19}\%\\

\hline
\hline
\end{tabular}}
\end{center}
\end{table*}


The experimental results are shown in Table~\ref{tab:abl}.
The results indicate that the impact of self-supervised adversarial training on the recognition of clean facial images is not significant.
However, the performance of the model without self-supervised adversarial training exhibited a significant decline in adversarial defense, especially under the strongest PGD attack.
This is because, if self-supervised adversarial training is not conducted, the model's training objective is face image reconstruction, and it will tend to incorporate all eigenvectors into the antibody to reconstruct all the details of the input image.
Although the regularization term in Eq.~\ref{eq:affinity} constrains the number of eigenvectors contained in the antibody to encourage the model to focus on important eigenvectors, the model's performance in adversarial defense is significantly reduced because it has not undergone targeted adversarial noise removal training.
%
%
These results also demonstrate the significant improvement in the defense capability of the model by self-supervised adversarial training.


\paragraph{Ablation Study for Memory Module:}
%
%
Drawing inspiration from the memory mechanisms of the immune system, a memory module is incorporated into the proposed defense model.
The memory module can store noise patterns during the training process and guide for generating antibodies through memory retrieval.
To verify the effectiveness of the memory module, we trained and tested the model without using the memory model and compared its defense performance with the original model.
When the memory module is not utilized, $\hat{f}_n$ in Eq.~\ref{eq:G} is simply substituted with $f_n$.
We conducted experiments using three adversarial attack methods, namely FGSM, DeepFool, and PGD, on the LFW dataset.
The testing protocol of the experiment is also consistent with Section~\ref{sec:ExpGeneral}.


The experimental results are shown in Table~\ref{tab:abl}.
The results indicate that when the memory module is not utilized, the proposed method performs similarly to the original model on clean facial images, but the adversarial defense performance decreases.
This is because the recognition performance on clean facial images only depends on the model's reconstruction ability, while the adversarial defense capability also requires the model to have strong noise removal ability.
The noise features stored in the memory module can provide guidance for generating antibodies, preventing the model from the scenario where fixing one vulnerability leads to the emergence of another.

\paragraph{Transferability to Different Face Recognition Models:}
%
In previous experiments, face recognition models that were compatible with the proposed defense method all utilized ArcFace (ResNet-50)~\cite{Deng2018ArcFace}.
The purpose of conducting this experiment is to answer the question of whether the proposed defense method can be implemented with other face recognition models for adversarial defense.
MobiFace~\cite{Duong2019Mobiface}, as a lightweight face recognition model, has many differences in design and structure compared to ArcFace.
MobiFace is utilized in tandem with the proposed defense model without undergoing retraining, but instead, directly integrated with it.
%
The experiments also employed FGSM, DeepFool, and PGD as adversarial attack methods, conducted on LFW with experimental settings consistent with Section~\ref{sec:ExpGeneral}.

\begin{table*}[t]
\begin{center}
\caption{Experimental results in conjunction with MobiFace (EER).}
\label{tab:ablMobi}
{
\begin{tabular}{c|c|c|c|c}
\hline
\hline
\bf{Attack Method} & \bf{Clean}$\downarrow$ & \bf{FGSM}$\downarrow$ & \bf{DeepFool} $\downarrow$ & \bf{PGD}$\downarrow$ \\
\hline
 MobiFace w/o defense    & 0.67\% & 57.72\% & 82.58\% & 99.19\%   \\
 MobiFace with defense    & 1.56\% & 7.37\% & 7.59\% & 24.82\%   \\
 \hline
 ArcFace w/o defense    & \bf{0.44}\% & 41.97\% & 89.49\% & 99.71\%   \\
 ArcFace with defense    & 1.01\% & \bf{4.46}\% & \bf{5.01}\% & \bf{14.19}\%   \\

\hline
\hline
\end{tabular}}
\end{center}
\end{table*}


The experimental results are shown in Table~\ref{tab:ablMobi}.
It can be observed that the proposed defense model performs worse than ArcFace when used in conjunction with MobiFace.
This is mainly due to two reasons:
Firstly, MobiFace itself is a lightweight model with weaker recognition ability than ArcFace, which can be demonstrated by the performance difference between them on clean facial images.
Secondly, the cooperation between ArcFace and the defense model is more harmonious, as ArcFace participates as the face recognition model in both antibody affinity measurement and adversarial self-supervised training.
However, even so, the proposed method still outperforms the compared methods in Table~\ref{tab:ICLFW}, which indicates its ability to be directly applied to other face recognition models without retraining.

\paragraph{Sensitivity Analysis on Hyper-parameters:}
%
There is a key hyper-parameter in Algorithm~\ref{alg:antibody}, which is the number of cloned antibodies $k$ at each iteration.
To investigate the effect of the number of cloned antibodies on the defense performance, we conducted comparative experiments with different values of $k$, namely $k=10$, $k=20$, and $k=30$.
Other than the number of cloned antibodies, the training settings for the three defense models were consistent with Section~\ref{sec:detail}.
FGSM, DeepFool, and PGD are utilized as adversarial attack methods for testing on LFW, with experimental settings consistent with Section~\ref{sec:ExpGeneral}.

\begin{table*}[t]
\begin{center}
\caption{Sensitivity analysis on the number of cloned antibodies $k$.}
\label{tab:ablsample}
{
\begin{tabular}{c|c|c|c|c}
\hline
\hline
\bf{Attack Method} & \bf{Clean}$\downarrow$ & \bf{FGSM}$\downarrow$ & \bf{DeepFool} $\downarrow$ & \bf{PGD}$\downarrow$ \\
\hline
 $k=10$    & \underline{1.01}\% & 4.46\%        & \underline{5.01}\%      & 14.19\%   \\
 \hline
 $k=20$    & 1.12\%        & \underline{4.43}\% & 5.13\%      & \underline{14.09}\%   \\
 \hline
 $k=30$    & \bf{0.87}\%   & \bf{4.36}\%   & \bf{4.85}\% & \bf{13.98}\%   \\

\hline
\hline
\end{tabular}}
\end{center}
\end{table*}


The experimental results are shown in Table~\ref{tab:ablsample}.
When $k=10$ and $k=20$, the model's performance is quite comparable.
When $k=30$, the model's performance is optimal, in terms of both recognition accuracy on clean facial images and its ability to defend against adversarial attacks.
In general, increasing the number of cloned antibodies can enhance the model's defense performance.
This is due to the fact that increasing the quantity of cloned antibodies enables the model to undertake a broader exploration during the optimization process, and also facilitates the exploitation of antibody mutation.
It is worth noting, however, that the performance improvement resulting from increasing the number of cloned antibodies is limited, and it also amplifies the computational complexity of model optimization.


\subsection{Failure Cases Analysis}



In this subsection, we delve into the analysis of the model's failure cases.
Such an examination is instrumental in gaining a more profound insight into the proposed model, significantly enhancing our understanding of its defensive performance.

\begin{figure}[h]
\begin{center}
\includegraphics[width=0.6\linewidth]{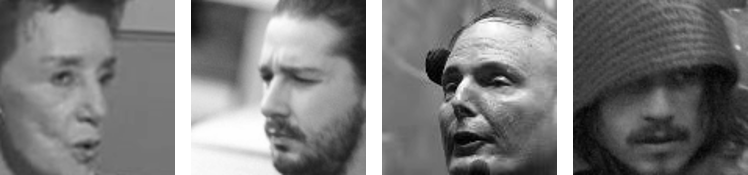}
\end{center}
   \caption{Examples of failure cases. The first three samples of defense failure exhibit significant head postures. The last sample is an occluded face image.}
\label{fig:failcase}
\end{figure}

Fig.~\ref{fig:failcase} displays four samples where adversarial defense failed.
It is evident from Fig.~\ref{fig:failcase} that the first three samples of defense failure exhibit significant head postures.
The reason for the model's failure on these samples lies in the composition of the antibodies.
The facial images with large head postures exhibit a substantial difference in data distribution compared to frontal face images.
Since the eigenvectors constituting the antibodies are derived from a data distribution predominantly composed of frontal face data (CelebA), their capacity to recover data with significant head postures, such as side faces, is limited.
This limitation adversely affects the defensive capabilities of the antibodies.

The fourth failure sample is an occluded face image.
Occlusions have a notable impact on face feature extraction, and the intra-class distance between clean images significantly increases due to occlusions.
This leads to samples, post adversarial noise removal, being challenging to recognize normally.

\section{Conclusion}
\label{sec:Conclusion}
%

In response to the challenges posed by the specificity and diversity of adversarial noises in face recognition, we draw inspiration from the working mechanism of the immune system and propose an adversarial defense method specifically designed for face recognition tasks in this paper.
Extensive experimental results demonstrate the efficacy of the proposed method, surpassing state-of-the-art adversarial defense methods.
Through a series of experiments and analyses, the advantages of the proposed method can be summarized as follows:
\begin{itemize}
    \item Regarding the strong specificity of adversarial noise in facial recognition, the proposed method can offer specific noise removal strategies for each input sample.
    This enables the effective removal of adversarial noises while preserving essential facial features.

    \item Through the proposed self-supervised adversarial training, the contradiction between the quantity and consistency of adversarial samples is resolved, thereby providing effective guidance for the optimization of defense models.

    \item The proposed method possesses dynamic adaptability, allowing it to autonomously adjust the optimization process of the defense model to accommodate changes in training data and objectives.
    This aspect holds inspiring implications for researchers to design novel methodologies based on this foundation.


    %


    \item The proposed method demonstrates a high level of applicability across various facial recognition models, as it can be directly employed without the need for retraining when applied to different facial recognition systems.

    \item The Artificial Immune System proposed in this paper also offers inspiration for other face-related security tasks, such as defense against face presentation attacks and DeepFakes attacks. Incorporating the principles of antibody cloning, mutation, selection, and memory mechanisms into these tasks could enhance the model's performance, particularly in terms of dynamic adaptability and generalization capabilities against various attack methodologies. This improvement presupposes the design of appropriate antibody forms and optimization objectives.

\end{itemize}




\section*{Data Avaliability Statement}

The data that support the findings of this study are available in Large-scale CelebFaces Attributes (CelebA) Dataset: https://mmlab.ie.cuhk.edu.hk/projects/CelebA.html, Labeled Faces in the Wild: http://vis-www.cs.umass.edu/lfw/, and MegaFace Dataset: http://megaface.cs.washington.edu/dataset/download.html.

\section*{Acknowledgement}
The authors would like to thank the associate editor and the reviewers for their valuable comments and advices.

This work is jointly supported by the National Key Research and Development Program of China (2022YFC3310400), the China Postdoctoral Science Foundation (BX20230044, 2023M730290), the National Natural Science Foundation of China (62276025, U23B2054, 62276263), and the Shenzhen Technology Plan Program (KQTD20170331093217368).


\bibliography{sn-article}

\end{document}